\documentclass[conference]{IEEEtran}
\IEEEoverridecommandlockouts
\usepackage{cite}
\usepackage{amsmath,amssymb,amsfonts}
\usepackage{graphicx}
\usepackage{textcomp}
\usepackage{xcolor}
\def\BibTeX{{\rm B\kern-.05em{\sc i\kern-.025em b}\kern-.08em
    T\kern-.1667em\lower.7ex\hbox{E}\kern-.125emX}}

\usepackage[utf8]{inputenc}
\usepackage{bm}
\usepackage{bbm}
\usepackage{dsfont}
\usepackage[ruled]{algorithm2e}
\usepackage{multirow}
\usepackage[T1]{fontenc}
\usepackage{wrapfig}
\usepackage{subfiles}
\usepackage{verbatim}
\usepackage{url}
\usepackage{seqsplit}
\usepackage{hyperref}
    
\newcommand{\dt}{cycle time}
\newcommand{\delt}{$\delta t$}
\newcommand{\bdt}{$b_{\delta t}$}
\newcommand{\mdt}{$m_{\delta t}$}
\newcommand{\gdt}{$\gamma_{\delta t}$}
\newcommand{\ldt}{$\lambda_{\delta t}$}
\newcommand{\argmax}{\mathop{\mathrm{argmax}}}
\newcommand{\argmin}{\mathop{\mathrm{argmin}}}

\begin{document}

\title{Reducing the Cost of Cycle-Time Tuning\\for Real-World Policy Optimization
\thanks{The authors thank the Reinforcement Learning and Artificial Intelligence Laboratory, Alberta Machine Intelligence Institute, Canada CIFAR AI Chairs Program, and Kindred Inc. for their generous support.}
}

\author{\IEEEauthorblockN{Homayoon Farrahi}
\IEEEauthorblockA{\textit{Department of Computing Science} \\
\textit{University of Alberta}\\
Edmonton, Canada \\
farrahi@ualberta.ca}
\and
\IEEEauthorblockN{A.\ Rupam Mahmood}
\IEEEauthorblockA{\textit{Department of Computing Science} \\
\textit{University of Alberta}\\
Edmonton, Canada \\
armahmood@ualberta.ca}
}

\maketitle

\begin{abstract}
Continuous-time reinforcement learning tasks commonly use discrete steps of fixed cycle times for actions.
As practitioners need to choose the action-cycle time for a given task, a significant concern is whether the hyper-parameters of the learning algorithm need to be re-tuned for each choice of the cycle time, which is prohibitive for real-world robotics.
In this work, we investigate the widely-used \textit{baseline} hyper-parameter values of two policy gradient algorithms---PPO and SAC---across different cycle times.
Using a benchmark task where the baseline hyper-parameters of both algorithms were shown to work well, we reveal that when a cycle time different than the task default is chosen, PPO with baseline hyper-parameters fails to learn.
Moreover, both PPO and SAC with their baseline hyper-parameters perform substantially worse than their tuned values for each cycle time.
We propose novel approaches for setting these hyper-parameters based on the cycle time.
In our experiments on simulated and real-world robotic tasks, the proposed approaches performed at least as well as the baseline hyper-parameters, with significantly better performance for most choices of the cycle time, and did not result in learning failure for any cycle time.
Hyper-parameter tuning still remains a significant barrier for real-world robotics, as our approaches require some initial tuning on a new task, even though it is negligible compared to an extensive tuning for each cycle time.
Our approach requires no additional tuning after the cycle time is changed for a given task and is a step toward avoiding extensive and costly hyper-parameter tuning for real-world policy optimization.
\end{abstract}

\section{Introduction}

Continuous control with policy gradient methods in deep reinforcement learning (RL) has emerged as a promising approach to robot learning.
However, except some recent works (Haarnoja et al.\ 2018b, Mahmood et al.\ 2018, Gupta et al.\ 2021), much of the advancements in this area rely on learning solely in simulations (Tan et al.\ 2018, Akkaya et al.\ 2019).
Learning to control directly using physical robots remains a challenge due to many barriers such as sample inefficiency, partial observability, and delays in a real-time environment (Dulac-Arnold et al.\ 2021).

An oft-ignored issue of real-world RL is that, in a new physical environment, a choice regarding time discretization has to be made explicitly, whereas in simulated benchmark environments this choice is already made.
In RL tasks, the continuous time is typically discretized into steps of an equal duration called the \textit{action-cycle time}, which refers to the time elapsed between two consecutive actions.
In benchmark simulated environments, such as OpenAI Gym (Brockman et al.\ 2016) and DeepMind Control Suite (Tassa et al.\ 2020), the action-cycle time is usually already chosen to a suitable value for learning, and the choice of action-cycle time is not exposed as a subject of study.
As a consequence, there is an absence of thorough investigations on how the existing algorithms perform across different cycle times.

When practitioners apply an RL algorithm to a new physical robotic task and need to tune the value of the action-cycle time for that task, a significant concern is how to choose the hyper-parameters of the algorithm.
Tuning hyper-parameters for real-world tasks is difficult but not unprecedented (Mahmood et al.~2018).
However, as a practitioner searches for a suitable cycle time, tuning the hyper-parameters for each choice of the cycle time becomes prohibitive.
It will be of substantial advantage if the values of the hyper-parameters could be chosen robustly for all cycle times or adjusted based on the cycle time without re-tuning.

In this work, we study the hyper-parameters of two policy gradient methods called \emph{Proximal Policy Optimization} (PPO, Schulman et al.\ 2017) and \emph{Soft Actor-Critic} (SAC, Haarnoja et al.\ 2018a) across different cycle times.
These two methods have been shown to perform effectively on many simulated benchmark tasks (Keng et al.\ 2019) and used for comparisons and further improvements in many other works (Fujimoto et al.\ 2018, Raffin \& Stulp 2020) without extensive hyper-parameter tuning.
In those works, roughly the same sets of hyper-parameter values are used separately for these two methods.
These baseline hyper-parameter values are from the original works introducing the methods, and similar values are also used as the default values in widely-available implementations.

Our experiments on a widely-used benchmark simulation show that the baseline hyper-parameter values learn effectively with the default cycle time of the environment. However, when the cycle time changes, the baseline values of PPO fail to learn.
Moreover, both PPO and SAC with their baseline values perform substantially worse than their tuned values for each cycle time.
We also find that their tuned hyper-parameter values are different for different cycle times.
Therefore, the only known approach left for cycle-time tuning is to re-tune the hyper-parameters for each cycle time.

To reduce the cost associated with cycle-time tuning, we propose new approaches for setting algorithm hyper-parameters that adapt to new cycle times without re-tuning (see also Farrahi 2021). Our proposed approaches perform at least as well as and for many cycle times substantially better than the baseline values. Hyper-parameter values set by our recommendations do not fail to learn at any cycle time we used whereas the baseline values do. Our approach requires an initial tuning to an arbitrary cycle time on a task to enable the transfer of hyper-parameter values to different cycle times on the same task. Although an initial tuning is still needed, it is an strict improvement over extensive tuning for every new cycle time. Our approach therefore reduces the cost associated with hyper-parameter tuning when cycle time changes. We validate our recommendations on two held-out simulated and real-world robotic tasks.
Our implementation of the tasks and experiments are publicly available at \url{https://github.com/homayoonfarrahi/cycle-time-study} to encourage future studies on cycle time.

\section{Related Work}

Reducing the cycle time can inhibit effective learning in action-value methods.
Baird (1994) illustrated that, at small \dt{}s, values of different actions in the same state get closer to each other, making learning the action-value function more sensitive to noise and function approximation error. They noted the collapse of the action-value function to the state-value function in continuous time and proposed the Advantage Updating algorithm.
Their approach
was later extended to deep \textit{Q}-learning methods by Tallec et al.\ (2019).
Policy gradient methods are also susceptible to degraded performance as \dt{} gets smaller. The variance of likelihood-ratio policy gradient estimates can explode as \dt{} goes toward zero as shown in an example by Munos (2006). They formulated a model-based policy gradient estimate
assuming knowledge of the model and the gradient of the reward with respect to the state, which is hard to satisfy in many practical tasks.
Based on the Hamilton-Jacobi-Bellman (HJB) equation, Doya (2000) derived the continuous-time TD error for learning the value function and extended the actor-critic method to continuous time.
Lee and Sutton (2021) developed the theory for applying policy iteration methods to continuous-time systems.

The time it takes for an agent to output an action after an observation---the action delay---
is particularly important in real-world robotics since, unlike simulations (Firoiu et al.\ 2018), real-world environments do not halt their progress as the agent calculates an action (Chen et al.\ 2021).
Travnik et al.\ (2018)
minimized action delay by reordering algorithmic steps. Dulac-Arnold et al.\ (2021) observed deteriorating performance with increasing action and observation delay.
Ramstedt and Pal (2019) introduced the Real-Time Markov decision process and Actor-Critic in which \dt{} should ideally be equal to the time for a forward pass of the policy, which could vary greatly for different policy architectures.
In this work, we focus on the cycle time issues and assume that the chosen \dt{}s always fit the forward pass for action calculations.
Dulac-Arnold et al.\ (2021) showed that increasing \dt{} hurts task performance when using fixed hyper-parameters. The same can be true when reducing \dt{} with fixed hyper-parameter values as we will show later.

\section{The Problem Setup}

\begin{algorithm}[t]
\DontPrintSemicolon
\SetAlgoNoLine

  $\Psi \doteq \mathrm{Initialize} ()$

  Retrieve from $\Psi$: action-cycle time \delt{}, environment time step ${\delta t}_\mathrm{env}$, learning period $U$, batch size $b$, parameterized policy $\pi_{\bm\theta}(a|s)$

  \BlankLine

  Initialize Buffer $B$ with capacity $b$

  Initialize $S_0 \sim d_0(\cdot)$

  $j \doteq 0$ \tcp*[l]{episode step}

  $k \doteq 0$ \tcp*[l]{agent step}

  \For{environment step $i=0, 1, 2, ...$} {

      \uIf{$j \bmod \delta t / {\delta t}_{\mathrm{env}} = 0$} {

          $\tilde{S}_k \doteq S_i$

          $\tilde{R}_{k+1} \doteq 0$

          Calculate action $\tilde{A}_k \sim \pi_{\bm\theta}(\cdot | S_i)$

      }

      Apply $\tilde{A}_k$ and observe $R_{i+1}, S'_{i+1}$

      $\tilde{R}_{k+1} \doteq \tilde{R}_{k+1} + R_{i+1}$

      $T_{i+1} \doteq \mathds{1}_{S'_{i+1} \mathrm{\;is\;terminal}}$

      \uIf{$j+1 \bmod \delta t / {\delta t}_{\mathrm{env}} = 0\ \mathrm{or}\ T_{i+1}=1$} {
          Store transaction in buffer $B_{k} = \left( \tilde{S}_k, \tilde{A}_k, \tilde{R}_{k+1}, S'_{i+1}, T_{i+1} \right)$

          \uIf{$k+1 \bmod U = 0$} {
              $\Psi \doteq \mathrm{Learn} (B, \Psi)$
          }

          $k \doteq k + 1$
      }

      $j \doteq j + 1$

      \uIf{$T_{i+1}=1$} {
          Sample $S_{i+1} \sim d_0(\cdot)$

          $j \doteq 0$
      }
      \uElse{
          $S_{i+1} \doteq S'_{i+1}$
      }

  }

  \caption{Agent-environment interaction loop for different \dt{}s}
  \label{alg:experience_collection_different_dts}
\end{algorithm}

\begin{figure*}[t]
    \centering
    \begin{minipage}[t]{.48\textwidth}
        \centering
        \includegraphics[keepaspectratio=true, width=1.0\textwidth]{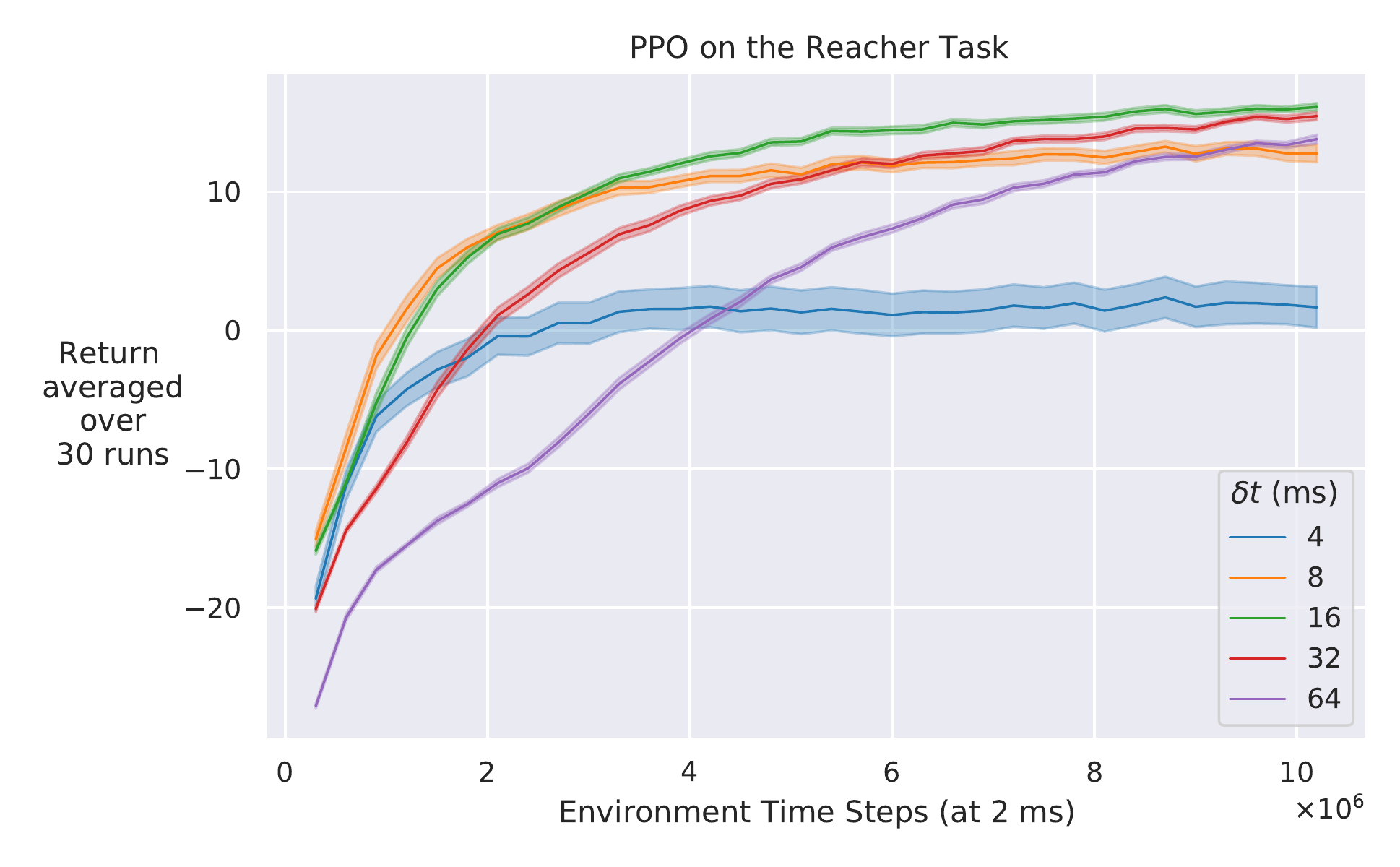}
        \caption {Learning curves for the baseline PPO hyper-parameters under different cycle times $\delta t$. Smaller $\delta t$s have worse asymptotic performance. Larger $\delta t$s learn more slowly.}
        \label{fig:defaultparams}
    \end{minipage}\hfill
    \begin{minipage}[t]{.48\textwidth}
        \centering
        \includegraphics[keepaspectratio=true, width=1.0\textwidth]{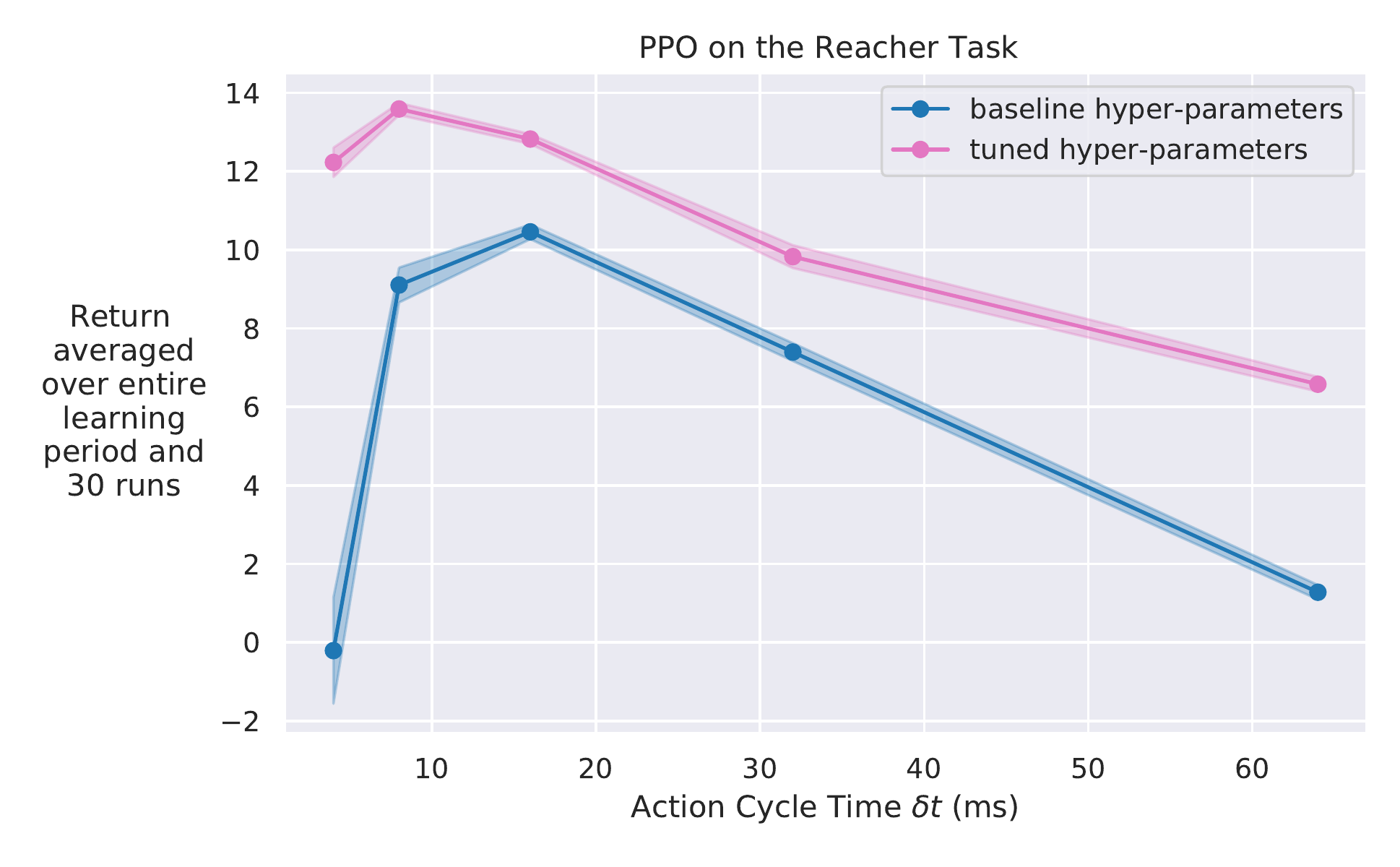}
    \caption {Performance vs.\ \dt{} $\delta t$ for different hyper-parameter configurations of PPO. Tuned hyper-parameters improved the learning performance substantially for all $\delta t$s.}
        \label{fig:auc_vs_dt_real}
    \end{minipage}
\end{figure*}

We use the undiscounted episodic reinforcement learning framework (Sutton \& Barto 2018), which is a special case of discounted episodic Markov Decision processes (MDPs).
In discounted episodic MDPs, an agent interacts with the environment through a sequence of episodes. Each episode starts with a state $S_0$ drawn from density $d_0$. At time step $t$, the agent takes action $A_t$ at state $S_t$ and receives next state $S_{t+1}$ and reward $R_{t+1}$ until termination.

In this framework, the episodic return is denoted by $G_0 \doteq \sum_{k=0}^{T-1} \gamma^k R_{k+1}$ with the discount factor $\gamma \in [0,1]$.
The value of a state $s$ under policy $\pi$ is defined as $\upsilon_\pi (s) \doteq \mathbb{E}_\pi \left[ G_0 \vert S_0=s \right]$.
When the policy $\pi$ is parameterized by $\bm\theta$, the goal of the agent is to change $\bm\theta$ to maximize the expected value of the initial state $J^\gamma (\bm\theta) \doteq \int_s d_0(s)\ \upsilon_{\pi_{\bm\theta}} (s)$.
In our problem formulation, $\gamma = 1$, and we seek to maximize the expected undiscounted episodic return $J^1 (\bm\theta)$, although solution methods often use a surrogate objective with $\gamma<1$.

Great care should be taken when running experiments with different \dt{}s in simulations.
The naive approach of changing the time interval of the underlying physics engine for simulated environments may lead to inconsistencies between different \dt{}s since the environment can behave differently over time even if it receives the same sequence of actions for both \dt{}s.
We instead run the environment at a small fixed environment time step ${\delta t}_\mathrm{env}$, and simulate other \dt{}s as integer multiples of ${\delta t}_\mathrm{env}$. Algorithm \ref{alg:experience_collection_different_dts} describes the agent-environment interaction loop for this setup.
This interaction loop is not specific to a particular learning algorithm, which can be instantiated by defining the \textit{Initialize} and \textit{Learn} functions as done in Appendix \ref{app:one-step-ac} for instance.

All parameters and hyper-parameters can be retrieved where needed from the set of all algorithm-specific parameters $\Psi$.
The agent-environment interaction happens every $\delta t / {\delta t}_\mathrm{env}$ environment steps, when the agent selects an action based on the latest environment state.
The selected action $\tilde{A}_k$ is repeatedly applied to the environment, and the received rewards are accumulated in $\tilde{R}_{k+1}$ until the next action selection.
Right before the next action selection, the transaction is stored in a buffer of size $b$, and the \textit{Learn} function is executed if $U$ agent-environment interactions have occurred.
The buffer overwrites its oldest sample in case it is full.
The two algorithms that we use for our experiments differ in their setting of batch size $b$ and learning period $U$.
In PPO (Appendix \ref{app:ppo_learn_function}), batch size $b$ is equal to learning period $U$, whereas SAC (Appendix \ref{app:sac_learn_function}) uses a much larger $b$ than PPO with $U=1$.
In each call of the \textit{Learn} function, PPO executes 10 epochs of mini-batch gradient updates, whereas SAC executes a single update.

\section{Studying the Baseline Hyper-Parameter Values of PPO}\label{sec:failure}

In this section, we investigate the baseline hyper-parameter values of PPO across different \dt{}s.
For our PPO experiments in simulated environments, we use baseline hyper-parameter values similar to the recommended Mujoco hyper-parameters in Schulman et al.\ (2017).
We use the PyBullet (Coumans \& Bai 2016) environment \textit{ReacherBulletEnv-v0} and refer to it as the \textit{Reacher Task}.
We modify its \textit{environment time step} from the default $16.5$ms to a constant $2$ms for all experiments, and the agent interacts with the environment at multiples of $2$ms for different cycle times.
To run an experiment with cycle time $\delta t=8$ms, for instance, the agent interacts with the environment every fourth environment step by taking the most recent observation as input and outputting an action, which is repeatedly applied to the environment until the next interaction.
Rewards are accumulated between two interactions, and all episodes last for $2.4$ seconds.
Some components of the reward function are scaled according to the environment time step to keep their relative weighting comparable to the original environment.

We ran PPO with the baseline hyper-parameter values
of Table \ref{tbl:pposimhps}
using various \dt{}s from $4$ms to $64$ms each for 10 million environment steps.
The learning curve for each \dt{} is given in Figure \ref{fig:defaultparams}. We further performed a grid search with five different values of batch size from 500 to 8000 and mini-batch size from 12 to 200.
For each set of hyper-parameters, average return over the entire learning period and $30$ independent runs was calculated, and the results for the best-performing tuned hyper-parameter values at each \dt{}
(Table \ref{tbl:pposimtunedhps})
were plotted in Figure \ref{fig:auc_vs_dt_real}.
We used separate runs to plot the results of the tuned hyper-parameter values to avoid maximization bias.
Similar curve of average returns with fixed baseline hyper-parameter values was also plotted in Figure \ref{fig:auc_vs_dt_real}.
For all plots in this paper, the results are calculated using undiscounted returns and shaded areas represent standard error.

Our experimental results in Figure \ref{fig:defaultparams} indicate that the baseline hyper-parameters can lead to learning failure when the cycle time changes, substantiated by the near zero return at 4ms, which does not give an effective goal-oriented behavior in this task.
Figure \ref{fig:defaultparams} shows that the asymptotic performance declines with small \dt{}s and that large \dt{}s hurt the learning speed.
The former may be because batches are collected more quickly, lacking enough information for useful updates.
The latter might be a result of fewer and infrequent updates.
Figure \ref{fig:auc_vs_dt_real} shows that tuning hyper-parameters to different values for each \dt{} leads to substantially increased performance compared to the baselines.
The tuned batch size and mini-batch size values are different from the baseline values and across cycle times.
This demonstrates the importance of having guidelines for adjusting different hyper-parameters based on \dt{} to allow transfer of hyper-parameters tuned to a specific \dt{} to a different \dt{} on the same task without re-tuning.

\section{Setting Hyper-Parameters of PPO as a Function of the Action-Cycle Time}

\begin{figure*}[t]
    \centering
    \begin{minipage}[t]{.48\textwidth}
        \centering
        \includegraphics[keepaspectratio=true, width=1.0\textwidth]{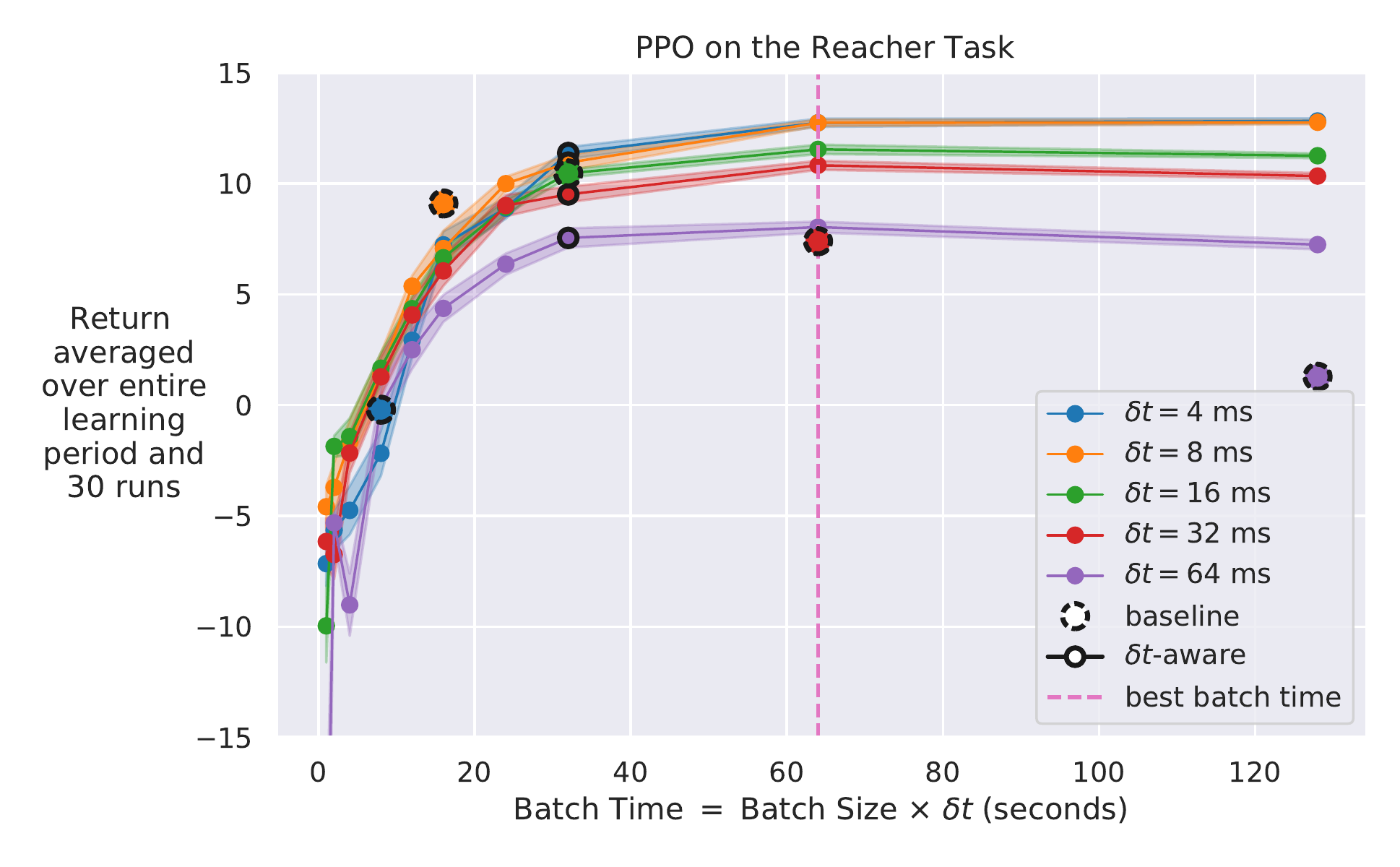}
        \caption {Performance using the \delt{}-aware hyper-parameters for different initial batch sizes.
        A close alignment of performance across different cycle times is obtained when batch time is kept constant.}
        \label{fig:auc_vs_bt_final}
    \end{minipage}\hfill
    \begin{minipage}[t]{.48\textwidth}
        \centering
        \includegraphics[keepaspectratio=true, width=1.0\textwidth]{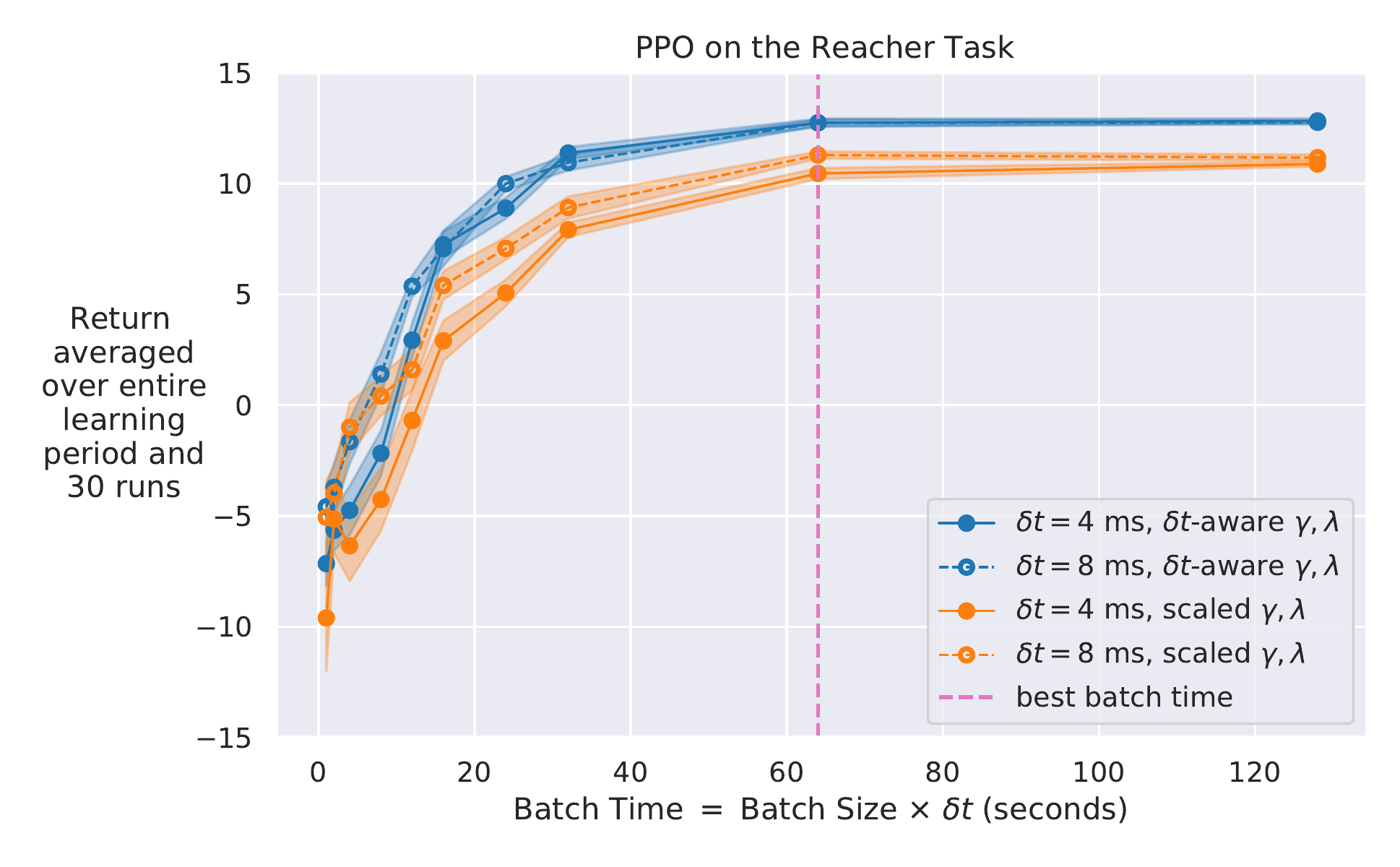}
        \caption {Performance of \delt{}-aware hyper-parameters for different initial batch sizes compared with ones where $\gamma$ and $\lambda$ are always exponentiated to the $\delta t / \delta t_0$ power (scaled).}
        \label{fig:auc_vs_bt_simple}
    \end{minipage}
\end{figure*}

Changing the cycle time may affect the relationship between time, and the hyper-parameters batch size $b$ and mini-batch size $m$, which could suggest the need for scaling them based on the \dt{}.
As the \dt{} decreases, each fixed-size batch or mini-batch of samples corresponds to less amount of real-time experience, possibly reducing the extent of useful information available in the batch or mini-batch.
To address this, we scale the baseline $b$ and $m$ inversely proportionally to the cycle time. Specifically, if $b_{\delta t_0}$ and $m_{\delta t_0}$ are the tuned or chosen batch size and mini-batch size, respectively, for the initial cycle time $\delta t_0$ (in ms), then the scaled batch size $b_{\delta t}$ and mini-batch size $m_{\delta t}$ for a new cycle time $\delta t$ are as follows: $b_{\delta t} \doteq \frac{{\delta t}_0}{\delta t} b_{{\delta t}_0},~~
    m_{\delta t} \doteq \frac{{\delta t}_0}{\delta t} m_{{\delta t}_0}$.
For instance, reducing \dt{} from the baseline ${\delta t}_0=16$ms to $8$ms, makes the batch size $b_8$ double the size of $b_{16}$.
This keeps the \textit{batch time} $\delta t \cdot b_{\delta t}$, the time it takes to collect a batch, and the amount of available information consistent across different \dt{}s.

The cycle time may influence the choice of the discount factor $\gamma$ and the trace-decay parameter $\lambda$ as well since it changes the rate at which rewards and $n$-step returns are discounted through time (Doya 2000, Tallec et al.\ 2019).
When using a small \dt{}, rewards and $n$-step returns are discounted more heavily for the same $\gamma$ and $\lambda$, as more experience samples are collected in a fixed time interval compared to larger \dt{}s.
Based on the above intuition discussed in previous works (Baird 1994, Doya 2000, Tallec et al.\ 2019), we exponentiated $\gamma$ and $\lambda$ to the ${\delta t}/{{\delta t}_0}$ power with initial cycle time $\delta t_0$.
However, this strategy was detrimental to the performance of smaller \dt{}s in our experiments perhaps since, as \dt{} gets smaller, an increasing number of samples are used to calculate the likelihood-ratio policy gradient estimate, possibly leading to its increased variance (Munos 2006). Hence, we only exponentiate $\gamma$ and $\lambda$ to the ${\delta t}/{{\delta t}_0}$ power for \dt{}s larger than $\delta t_0$.
This can be achieved by setting \gdt{} and \ldt{} to be the minimum of the baseline and the scaled one.
Based on the mentioned modifications, we present the new \textit{$\delta t$-aware hyper-parameters} \bdt{}, \mdt{}, \gdt{} and \ldt{}, which adapt to different \dt{}s as a function of the baseline action-cycle time ${\delta t}_0$ as follows:
$b_{\delta t} \doteq \frac{{\delta t}_0}{\delta t} b_{{\delta t}_0}$,
$m_{\delta t} \doteq \frac{{\delta t}_0}{\delta t} m_{{\delta t}_0}$,
$\gamma_{\delta t} \doteq \min\left(\gamma_{{\delta t}_0}, \gamma_{{\delta t}_0}^{{\delta t}/{{\delta t}_0}}\right)$, and
$\lambda_{\delta t} \doteq \min\left(\lambda_{{\delta t}_0}, \lambda_{{\delta t}_0}^{{\delta t}/{{\delta t}_0}}\right)$.
This proposed approach requires starting from a roughly tuned set of hyper-parameter values, which we call \emph{the initial hyper-parameter values}, at an initially chosen cycle time $\delta t_0$.

\begin{figure*}[t]
    \centering
    \begin{minipage}[t]{.48\textwidth}
        \centering
        \includegraphics[keepaspectratio=true, width=1.0\textwidth]{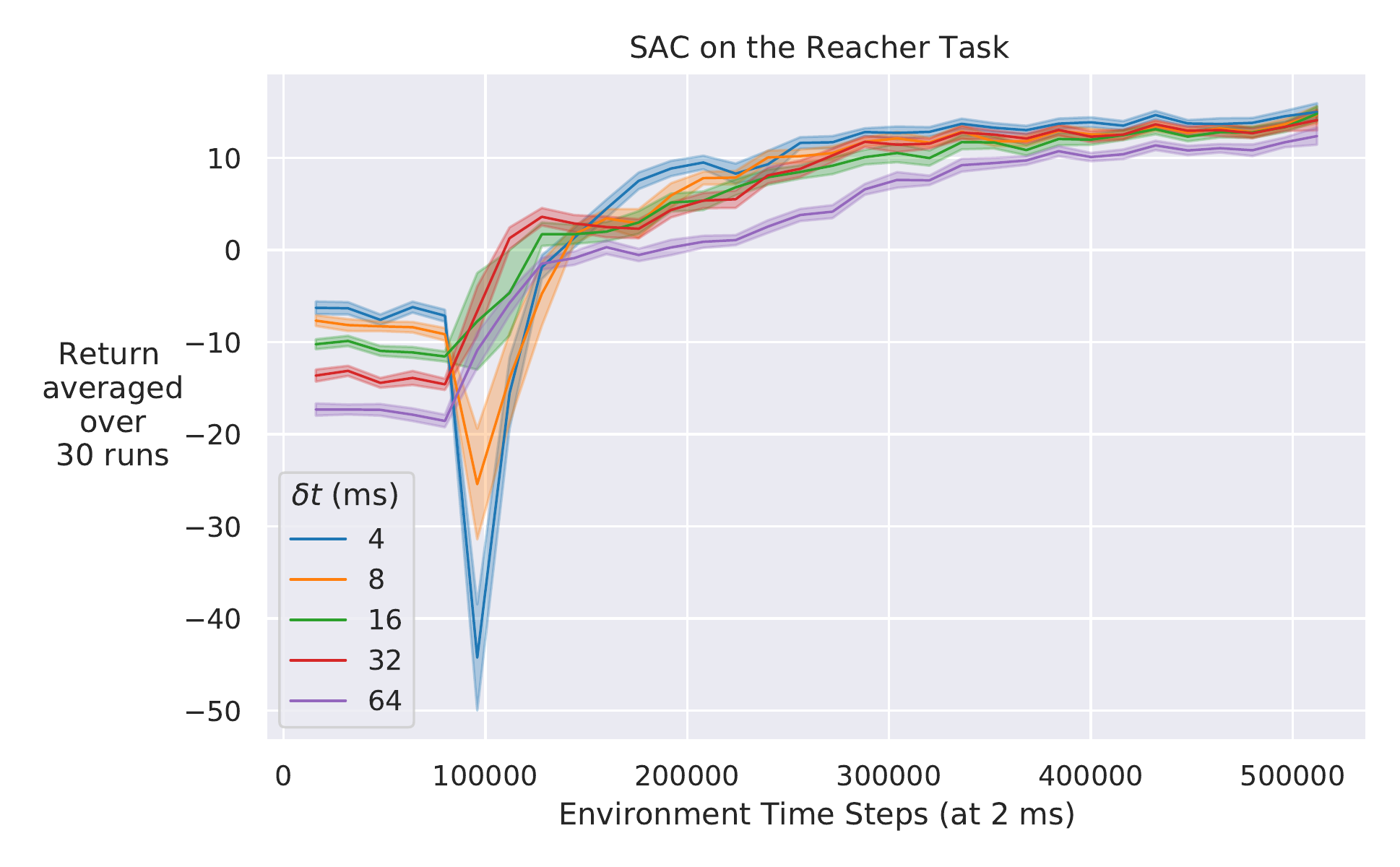}
        \caption {Learning curves of baseline SAC hyper-parameters.
        Smaller \dt{}s suffer a sharp decline in the beginning, and the largest \dt{} learns more slowly.
        }
        \label{fig:sac_defaultparams_reacher}
    \end{minipage}\hfill
    \begin{minipage}[t]{.48\textwidth}
        \centering
        \includegraphics[keepaspectratio=true, width=1.0\textwidth]{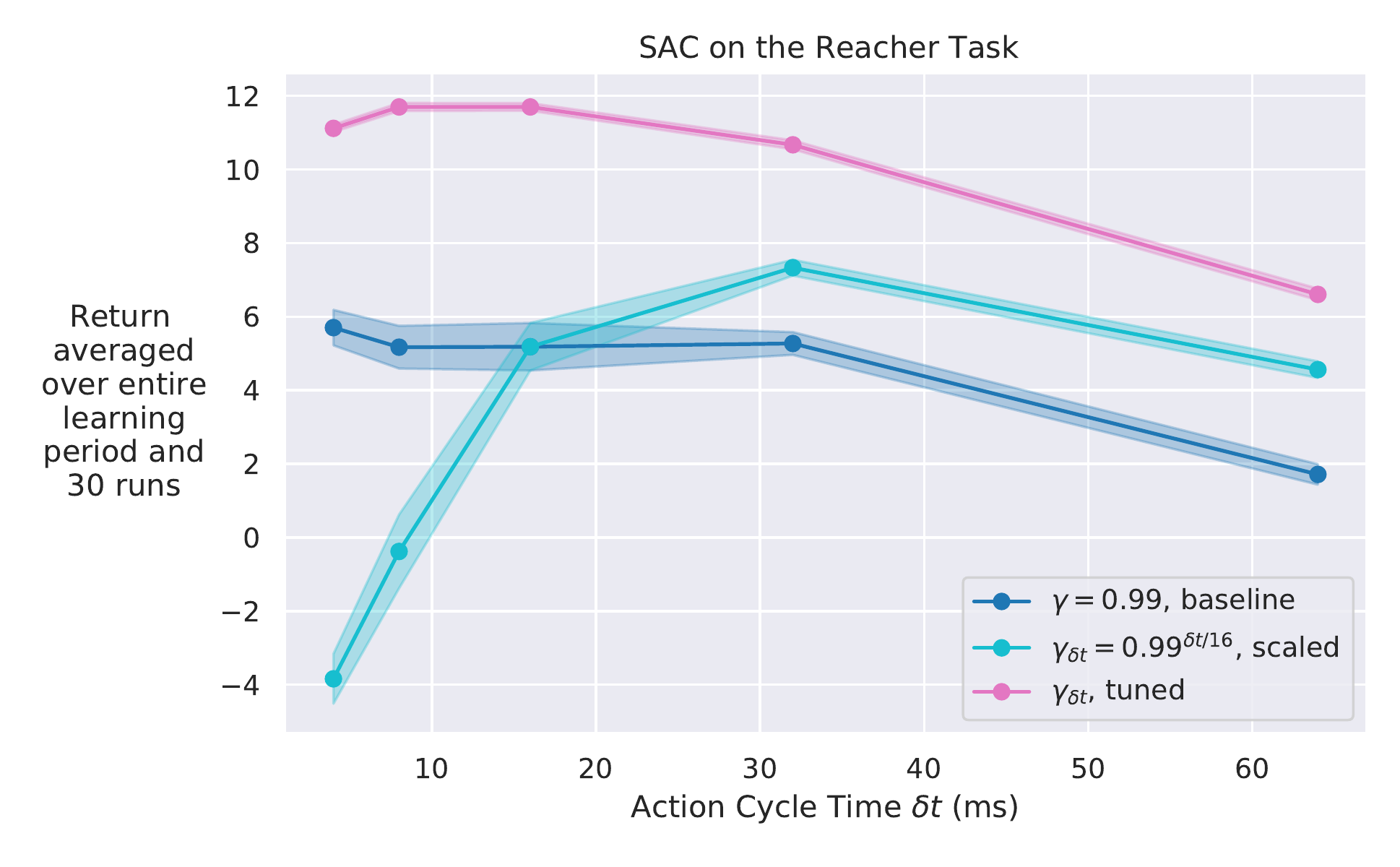}
        \caption {Performance vs.\ \dt{} for different choices of $\gamma$. Scaling $\gamma$ based on \dt{} makes the performance more sensitive to \dt{}. Performance can be improved by tuning $\gamma$ for each \dt{} separately.
        }
        \label{fig:sac_oar_vs_dt_reacher}
    \end{minipage}
\end{figure*}

We evaluated the \delt{}-aware hyper-parameters on the Reacher Task by setting the initial cycle time ${\delta t}_0=16$ms.
The initial values of $m_{\delta t_0}, \gamma_{\delta t_0}$, and $\lambda_{\delta t_0}$ were set to the baseline values $50$, $0.99$, and $0.95$ (Table \ref{tbl:pposimhps_dtaware}), respectively.
We varied the initial batch size $b_{\delta t_0}$, which gives different initial hyper-parameter values, the overall performance of which for ${\delta t}_0=16$ms is given by the green curve in Figure \ref{fig:auc_vs_bt_final}.
For each choice of the initial hyper-parameter values, we calculated the \delt{}-aware hyper-parameter values at different cycle times between 4ms and 64ms, and measured their overall performance, which are also given in Figure \ref{fig:auc_vs_bt_final} with curves of different colors for different cycle times.
With $\delta t$-aware hyper-parameter values, the curves aligned with each other well when the x-axis represents the batch time or the time it takes in seconds to collect a batch.
A benefit of this alignment is that, if we tune the batch size at any initial cycle time, then for other cycle times the $\delta t$-aware hyper-parameters would give a batch size that performs the best within that cycle time, indicated by the vertical purple line, which signifies keeping the batch time fixed at the tuned initial value when changing the cycle time.
For all \dt{}s, the \delt{}-aware hyper-parameters calculated from PPO baselines (small solid-bordered circles at 32s batch time), performed better than the PPO baselines (dash-bordered circles).
We also experimented with scaled batch and mini-batch sizes but $\gamma$ and $\lambda$ scaled without clipping.
As shown in Figure \ref{fig:auc_vs_bt_simple}, exponentiating $\gamma$ and $\lambda$ to the $\delta t / \delta t_0$ power reduces the performance of smaller cycle times for almost all batch times.
Other cycle times looked exactly the same as in Figure \ref{fig:auc_vs_bt_final} and were not drawn.
When $\gamma$ and $\lambda$ were instead kept constant at the baseline values for all cycle times, larger cycle times performed worse (Figure \ref{fig:auc_vs_bt_noscale_simple}).

\section{Setting Hyper-Parameters of SAC as a Function of the Action-Cycle Time}

In this section, we first examine how changing the cycle time influences the performance of SAC and then how scaling the discount factor $\gamma$ based on \dt{} affects the performance of different \dt{}s.
On the Reacher Task, we ran SAC
with the baseline hyper-parameter values given by Haarnoja et al.~(2018a) for 500,000 environment steps, and all other experimental details remain as in the previous sections. Figure \ref{fig:sac_defaultparams_reacher} shows the resulting learning curves for each \dt{}, averaged over $30$ independent runs.
As evident, the smaller cycle times 4ms and 8ms undergo sharp drops in performance just as the agent starts using the learned policy, although they are able to recover and match the performance of other \dt{}s swiftly. The larger $\delta t=64$ms learns more slowly throughout and never quite exceeds the learned performance of any other \dt{}.

We explained previously how changing the \dt{} can cause the rewards to be discounted faster or slower through time if $\gamma$ is kept constant.
As such, we continued by studying the effect of scaling and tuning the discount factor $\gamma$ on performance. We scaled the discount factor for each \dt{} according to $\gamma_{\delta t}=\gamma_{16}^{{\delta t}/16}$ with baseline $\gamma_{16}=0.99$ to ensure consistent discounting of the rewards through time (Doya 2000, Tallec et al.\ 2019).
Average returns over the entire learning period and $30$ runs were plotted in Figure \ref{fig:sac_oar_vs_dt_reacher} (curve in cyan). Similar curves were drawn for two additional sets of experiments.
In the first set, baseline hyper-parameter values including $\gamma=0.99$ stayed constant across all \dt{}s (curve in blue). In the second set, the best $\gamma_{\delta t}$ was found by searching over $12$ different values from $0.99^{128}=0.2763$ to $1.0$ for each \dt{} individually and rerun to avoid maximization bias (curve in purple).
Exponentiating the baseline $\gamma$ to the $\delta t/16$ power surprisingly makes SAC more sensitive to \dt{}, as it boosts the performance of larger \dt{}s and hinders that of smaller ones. Tuning $\gamma$ to different values, however, can substantially improve the results of all \dt{}s over the baseline and scaled values.
The tuned $\gamma$ values
(Table \ref{tbl:sacsimtunedhps})
are different from the baseline $\gamma=0.99$ and across cycle times.
Scaling $\gamma$ was not a better choice than the baseline, and tuning $\gamma$ outperformed both.

\begin{figure}[t]
    \centering
    \includegraphics[keepaspectratio=true, width=0.48\textwidth]{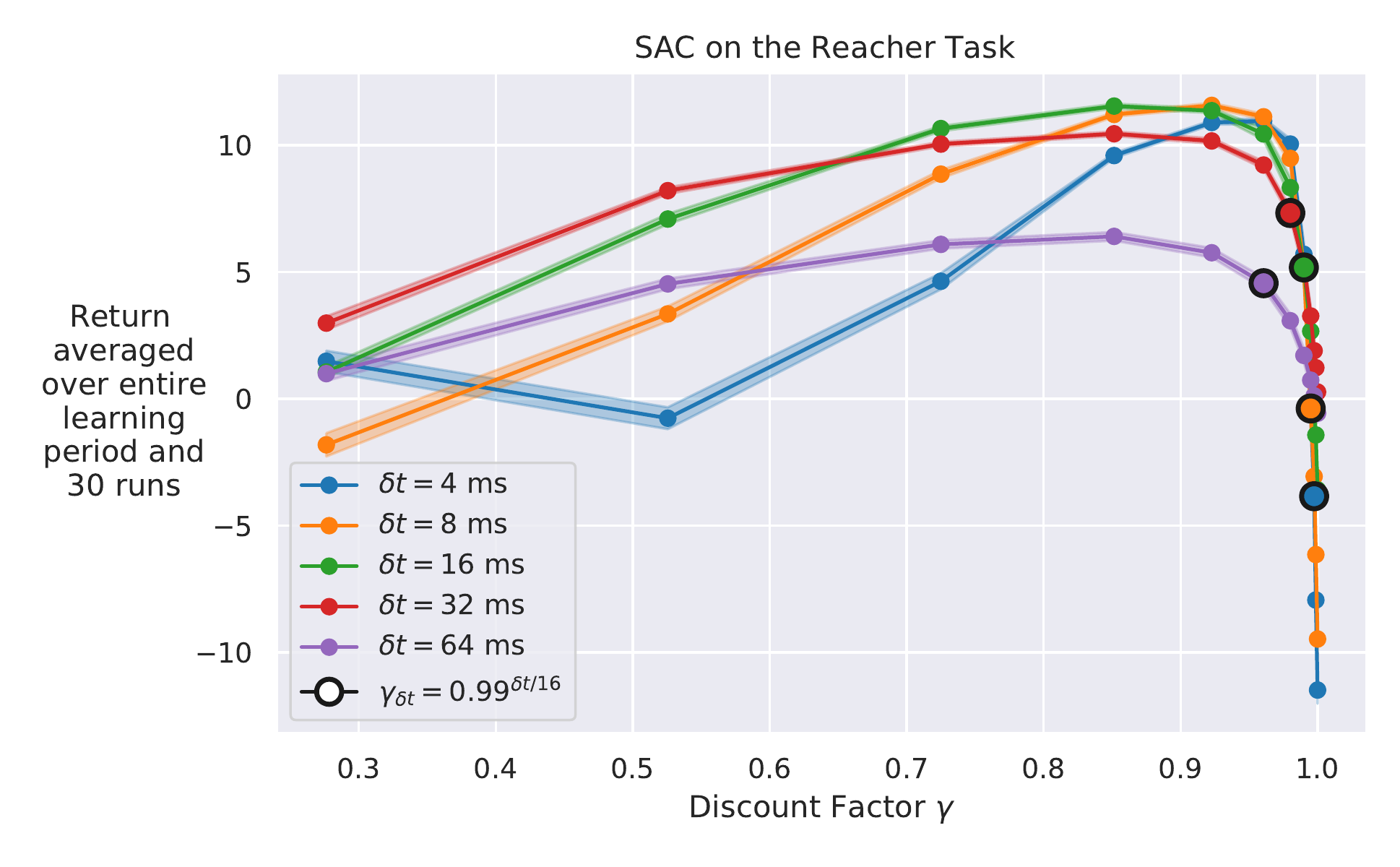}
    \caption {Performance for a sweep of $\gamma$ values.
    Peak performance is achieved at intermediate values of $\gamma$. Curves are stretched horizontally for increasing \dt{}, suggesting the need for scaling $\gamma$ based on \dt{}.
    }
    \label{fig:sac_oar_vs_gam_256_reacher}
\end{figure}

Figure \ref{fig:sac_oar_vs_gam_256_reacher} shows the average return of different $\gamma$s and cycle times to better illustrate the relationship between them.
The best performance at $\delta t=16$ms is reached with $\gamma \approx 0.851$, which is considerably far from the baseline value of $0.99$.
The performance of all \dt{}s peak at different intermediate values of $\gamma$ and steadily decline for both increasing and decreasing $\gamma$s (except for the curious jump of $\delta t=4$ms at the smallest $\gamma$).
The location of these peaks might be different for other tasks with some tasks obtaining the maximum performance with $\gamma$s larger than the baseline $0.99$.
In this task, the baseline $\gamma$ is larger than the best, and scaling it according to \dt{} hurts the performance of smaller \dt{}s (solid-borderd circles in Figure \ref{fig:sac_oar_vs_gam_256_reacher}). If the best $\gamma$ was larger than the baseline instead, scaling it would have helped smaller \dt{}s. This conflict points to the insufficiency of merely scaling $\gamma$ for adapting it to different \dt{}s.
In addition, as \dt{} gets larger, the curves are consistently stretched horizontally, which hints at the need for scaling $\gamma$ based on \dt{} to make the curves more aligned.

\begin{figure*}
    \centering
    \includegraphics[keepaspectratio=true, width=1.0\textwidth]{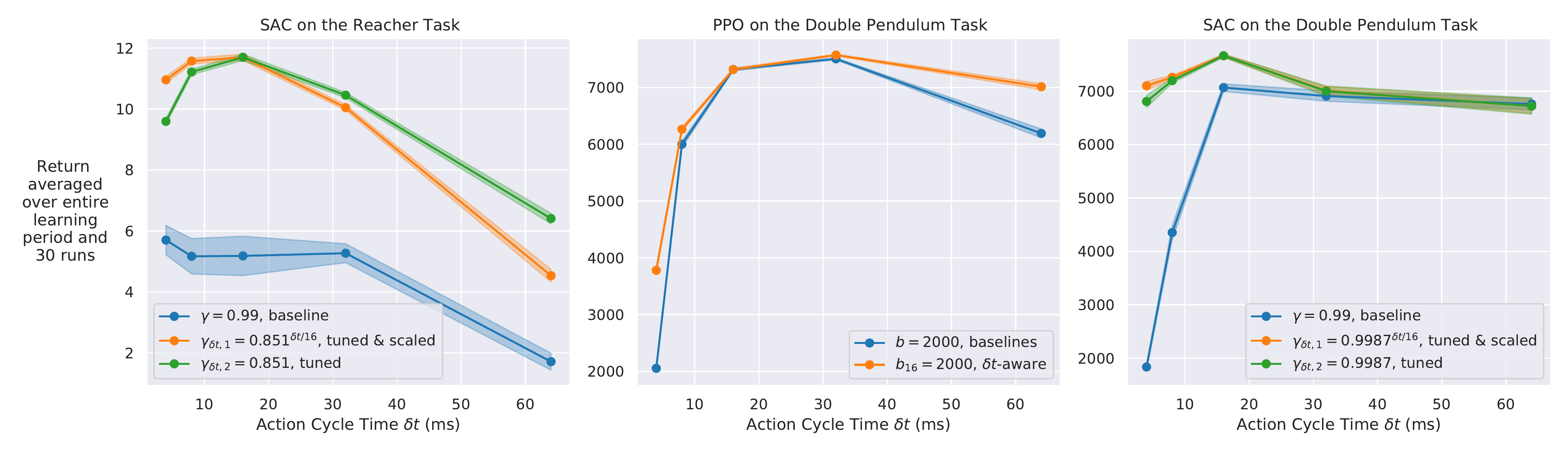}
    \caption {(left) Performance of the two hypotheses compared with baseline $\gamma$.
        Both offer notable performance gains over the baseline. Scaling $\gamma$ according to \dt{} favors smaller \dt{}s and hurts larger ones. (middle) Performance of \delt{}-aware and baseline hyper-parameters compared.
        The \delt{}-aware values improve over the baselines and do not fail to learn at $\delta t=4$ms. (right) Performance of \delt{}-aware \gdt{} of SAC compared with baseline $\gamma$.
        The tuned $\gamma_{\delta t, 2}$ fully regains the performance of smaller \dt{}s. Additional scaling ($\gamma_{\delta t, 1}$) causes only a marginal refinement.
    }
    \label{fig:icra2022_oar_vs_dt_combo}
\end{figure*}

Based on the results of the exhaustive $\gamma$ sweep, we propose two approaches for adapting $\gamma$ to different \dt{}s. In the first approach, $\gamma$ is initially tuned to find the best value at an initial ${\delta t}_0$ and subsequently scaled based on the \dt{}. In the second approach, the tuned $\gamma$ is kept constant for other \dt{}s. These two approaches are the new \textit{\delt{}-aware discount factor} $\gamma_{\delta t, 1}$ and $\gamma_{\delta t, 2}$ of SAC defined as
$\gamma_{\delta t, 1} \doteq (\argmax_{\gamma_{{\delta t}_0}} \hat{G_0})^{{\delta t}/{{\delta t}_0}}$
$\gamma_{\delta t, 2} \doteq (\argmax_{\gamma_{{\delta t}_0}} \hat{G_0})$
with ${\delta t}_0$ as the initial cycle time and $\hat{G_0}$ as the estimate of the undiscounted episodic return obtained as a sample average over the entire learning process.
We tested the two approaches for the \delt{}-aware \gdt{} on the Reacher Task with ${\delta t}_0=16$ms, for which $\gamma_{16} \approx 0.851$ performed best.
Figure \ref{fig:icra2022_oar_vs_dt_combo} (left) shows that both of these approaches performed significantly better than the baseline $\gamma=0.99$. It is thus clear that tuning $\gamma$ for SAC at ${\delta t}_0$ is a crucial step in adapting it to different \dt{}s. The additional scaling in $\gamma_{\delta t, 1}$ benefits smaller \dt{}s and hurts larger ones.

\section{Validating the \texorpdfstring{$\delta t$}{dt}-Aware Hyper-Parameters on a Held-Out Simulated Task}

In this section, we validate our proposed \delt{}-aware hyper-parameters on the PyBullet environment \textit{InvertedDoublePendulumBulletEnv-v0} referred to as the \textit{Double Pendulum Task}.
We reduced the environment time step from $16.5$ms to a constant $4$ms.
Episodes lasted a maximum of $16$ simulation seconds.
We ran two sets of experiments with PPO. For the first set, we kept the baseline hyper-parameters ($b=2000$) constant across \dt{}s, and for the second set, we used the \delt{}-aware hyper-parameters
of Table \ref{tbl:pposimhps_dtaware}
with ${\delta t}_0=16$ms ($b_{16}=2000$) and initial values set to the baselines. All runs lasted for $10$ million environment steps, and the average returns over the full learning period were plotted in Figure \ref{fig:icra2022_oar_vs_dt_combo} (middle).
We investigated the validity of both approaches for \gdt{} of SAC on the Double Pendulum Task and compared them with the baseline $\gamma=0.99$. Peak performance at ${\delta t}_0=16$ms was achieved with $\gamma=0.9987$, which was both held constant across and scaled based on different \dt{}s.
All runs lasted for one million environment steps with results depicted in Figure \ref{fig:icra2022_oar_vs_dt_combo} (right).
Other experimental details were similar to previous sections.

For both PPO and SAC, the \delt{}-aware hyper-parameters were an improvement over the baseline values and did not fail to learn at any cycle time, as opposed to the ineffective performance of baseline values at $\delta t=4$ms, which was confirmed by rendering the behavior after learning.
Figure \ref{fig:icra2022_oar_vs_dt_combo} (right) validates \gdt{} of SAC, as the requisite tuning of $\gamma$ done by $\gamma_{\delta t, 2}$ at an initial ${\delta t}_0$ retrieved the lost performance at smaller \dt{}s with only marginal refinement resulting from the subsequent scaling done by $\gamma_{\delta t, 1}$.
Contrary to the baselines, the \delt{}-aware hyper-parameters did not fail to learn at any cycle time.
The baseline $\gamma=0.99$ particularly impairs the performance of $\delta t=4$ and $\delta t=8$ms, which contrasts the outcomes from the Reacher Task shown in Figure \ref{fig:icra2022_oar_vs_dt_combo} (left).
The baseline value of $\gamma$ in the SAC algorithm, therefore, may not be effective across different environments and \dt{}s, which remains an issue for future studies.

\begin{figure}[t]
    \centering
    \includegraphics[keepaspectratio=true, width=0.35\textwidth]{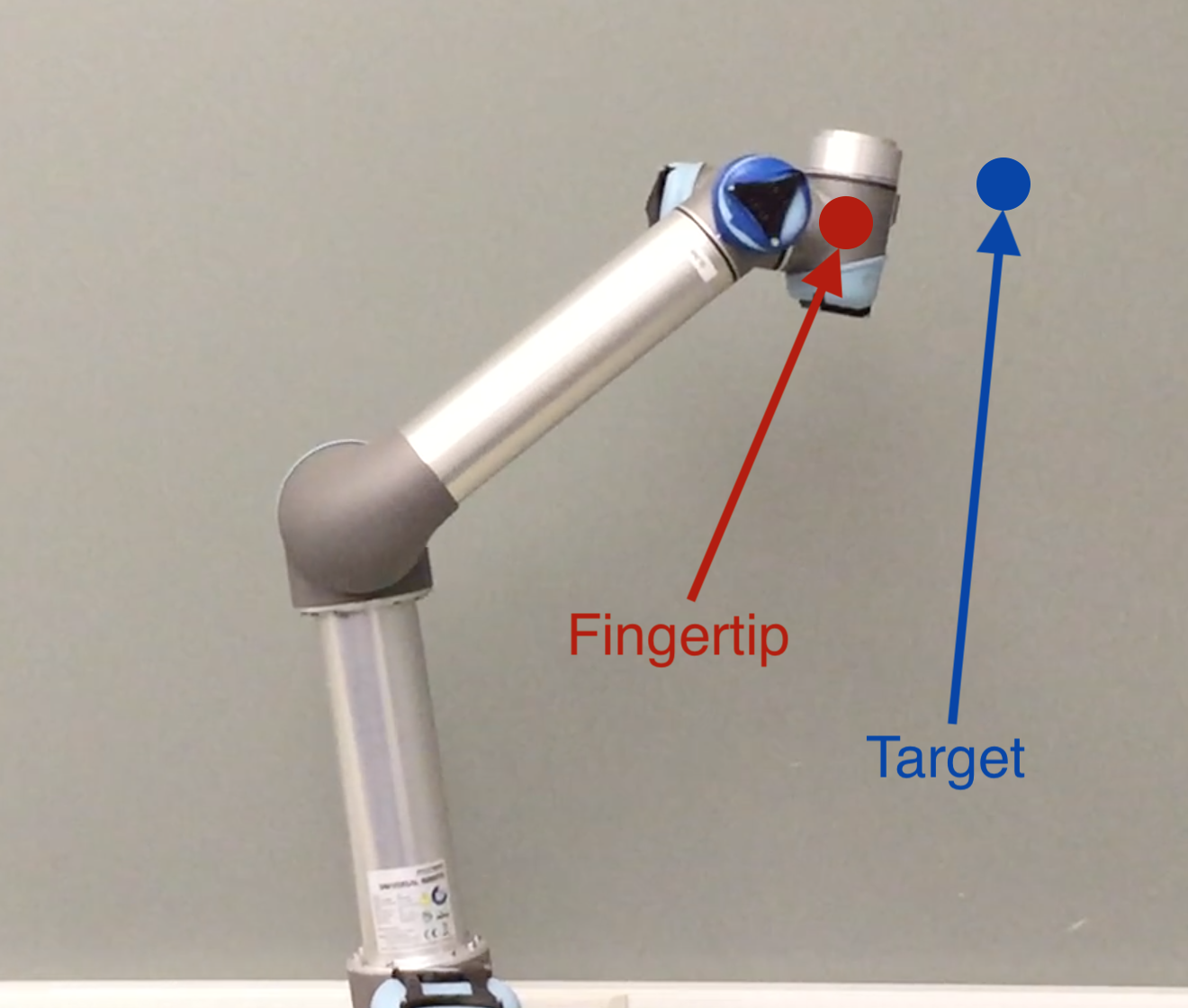}
    \caption {Real-Robot Reacher Task. The goal is to move the base and elbow joints to get the fingertip as close to target as fast as possible.}
    \label{fig:ur5}
\end{figure}

\section{Validation on a Real-World Robotic Task}

\begin{figure*}[t]
    \centering
    \begin{minipage}[t]{.48\textwidth}
        \centering
        \includegraphics[keepaspectratio=true, width=1.0\textwidth]{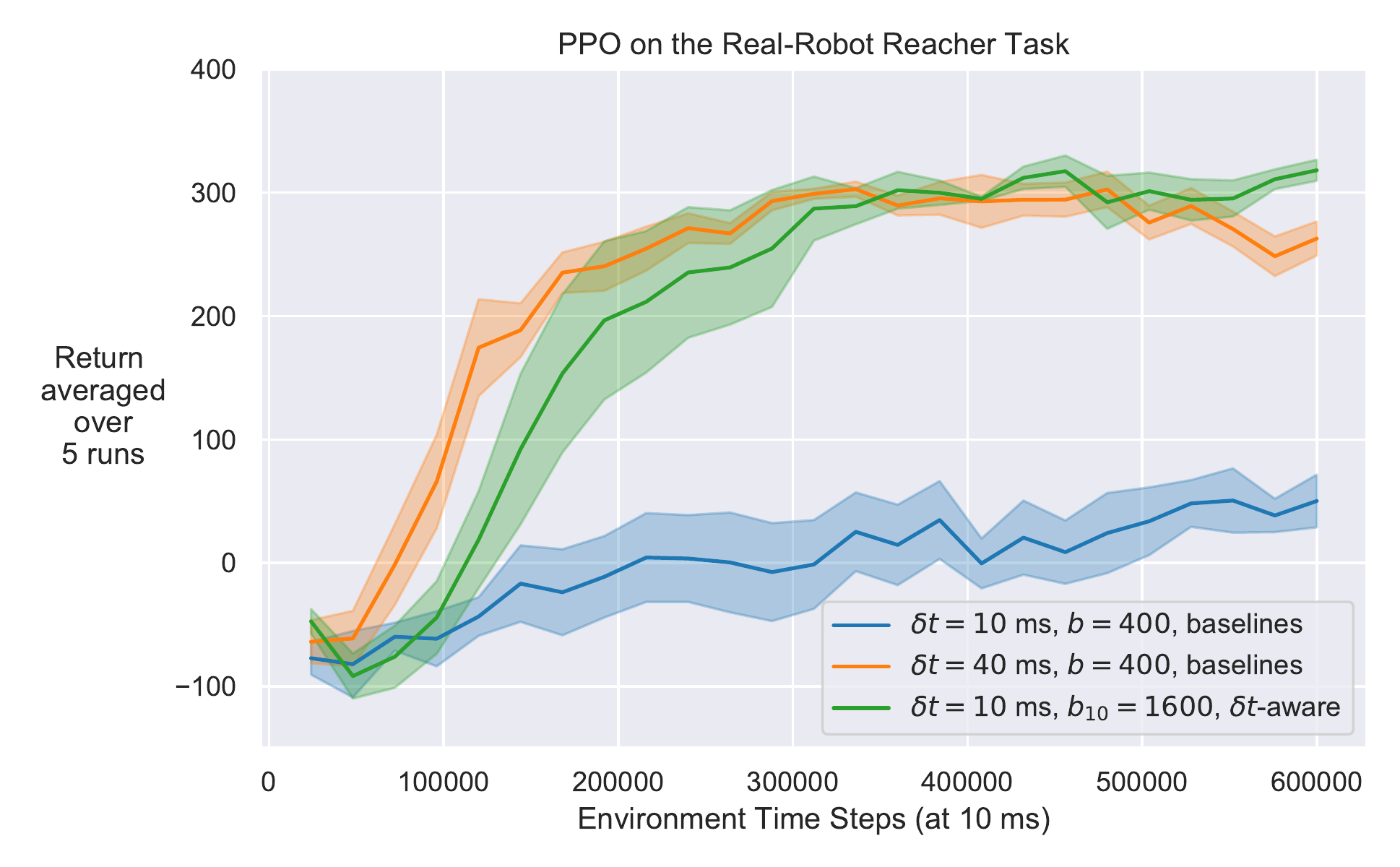}
        \caption {Learning curves of PPO on the Real-Robot Reacher Task comparing the \delt{}-aware hyper-parameters with the baseline ones. Asymptotic performance is recovered by the \delt{}-aware hyper-parameters.}
        \label{fig:ur5_learning_curves_4}
    \end{minipage}\hfill
    \begin{minipage}[t]{.48\textwidth}
        \centering
        \includegraphics[keepaspectratio=true, width=1.0\textwidth]{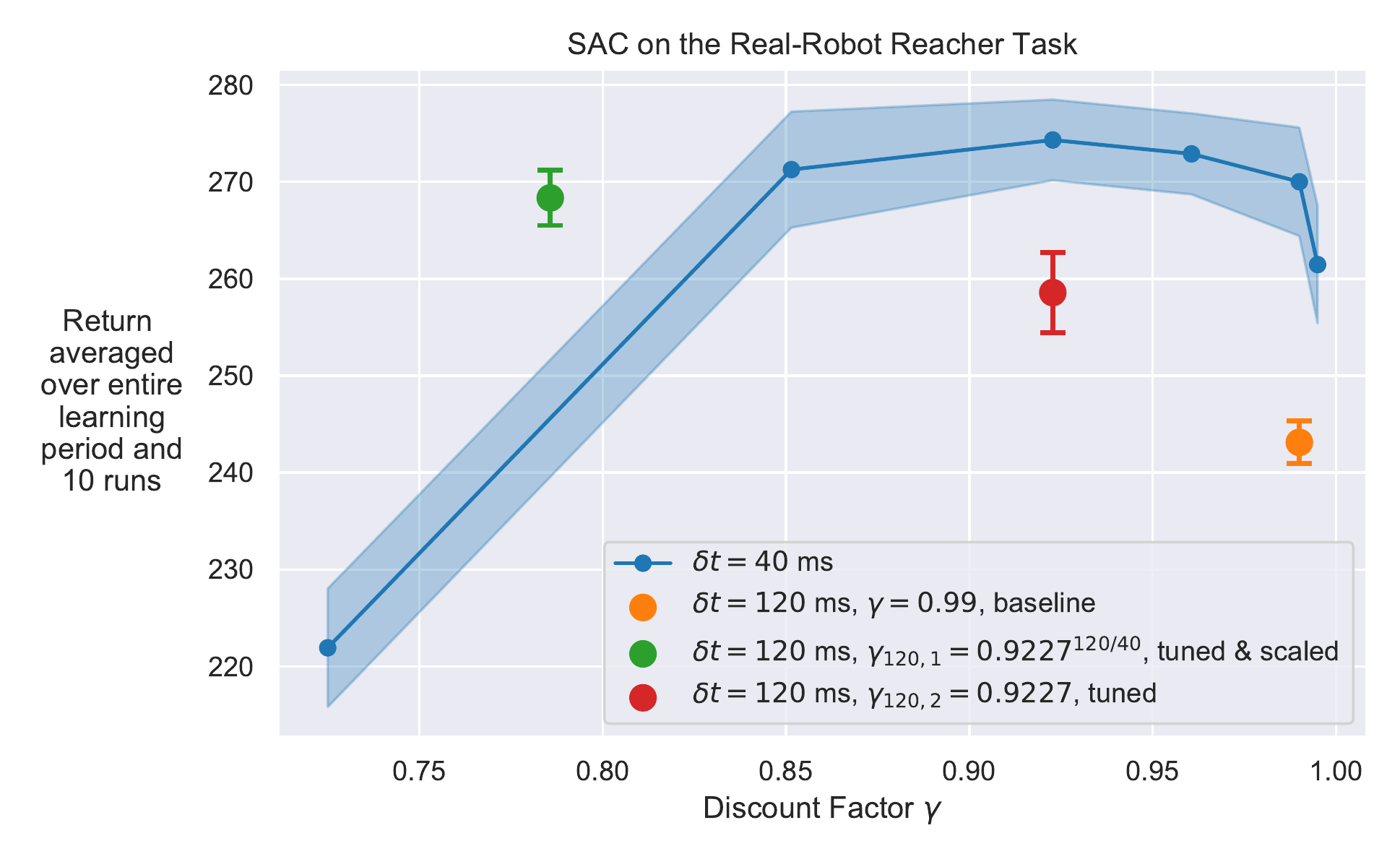}
        \caption {Performance of SAC comparing the baseline $\gamma$ with our two hypotheses on Real-Robot Reacher. The tuned and scaled $\gamma_{\delta t, 1}$ almost fully recovers the poor performance of the baseline $\gamma$ on $\delta t=120$ms.}
        \label{fig:sac_oar_vs_gam_256_urreacher}
    \end{minipage}
\end{figure*}

We validated the \delt{}-aware hyper-parameters of PPO and SAC on the \textit{UR-Reacher-2} task shown in Figure \ref{fig:ur5} developed by Mahmood et al.\ (2018), which we call the \textit{Real-Robot Reacher Task}. We
set the environment time step to $10$ms. Three sets of experiments were run using PPO. In the first set, we used ${\delta t}_0 = 40$ms as the benchmark. The other two sets used $\delta t = 10$ms to compare baseline initial hyper-parameter values
($b=400$, $m=10$, $\gamma=0.99$, $\lambda=0.95$), kept constant across \dt{}s, to the \delt{}-aware hyper-parameters that adapt to $\delta t=10$ms, e.g., $b_{10} = \frac{{\delta t}_0}{10}  b_{{\delta t}_0} = \frac{40}{10}\times 400 = 1600$.
The baseline values were chosen since they perform well on this task at ${\delta t}_0 = 40$.
All episodes were $4$ seconds long for all \dt{}s.
Each run lasted for 600,000 environment steps or 100 real-time minutes, and the learning curves, averaged over five runs, were plotted in Figure \ref{fig:ur5_learning_curves_4}.
For SAC, the environment time step was set to $40$ms to ensure the learning updates
take no longer than a cycle time. A $\gamma \approx 0.9227$ produced peak performance at ${\delta t}_0=40$ms. Three sets of experiments at $\delta t=120$ms were performed
using the baseline $\gamma=0.99$, the tuned $\gamma_{\delta t, 2} \approx 0.9227$, and the tuned and scaled $\gamma_{\delta t, 1} \approx 0.9227^{120/40} = 0.786$.
Figure \ref{fig:sac_oar_vs_gam_256_urreacher} shows the returns averaged over $10$ independent runs and 50,000 environment steps or 33 real-time minutes for each set of experiments.
The learning curves for $\delta t=40$ms and $\delta t=120$ms are in Figures \ref{fig:sac_oar_vs_timesteps_256_urreacher_baseline_learning_curves} and \ref{fig:sac_oar_vs_timesteps_256_urreacher_120ms}, respectively.
Videos of the learned policies can be seen at \url{https://www.youtube.com/watch?v=tmo5fWGRPtk}.

As seen in Figure \ref{fig:ur5_learning_curves_4}, the \delt{}-aware hyper-parameters of PPO on Real-Robot Reacher recovered the lost performance of baseline hyper-parameter values with a small \dt{}.
The \delt{}-aware hyper-parameters of PPO at $\delta t=10$ms achieved an asymptotic performance similar to or slightly better than the baselines at ${\delta t}_0=40$ms, whereas the baselines performed significantly worse at $\delta t=10$ms.
The performances of $\delta t=10$ms and ${\delta t}_0=40$ms were comparable with the Mujoco baseline values ($b=2000$) of Table \ref{tbl:pposimhps_dtaware} (Figure \ref{fig:ur5_learning_curves_2000}).
For the SAC experiments in Figure \ref{fig:sac_oar_vs_gam_256_urreacher}, the baseline $\gamma=0.99$ reduced the performance when \dt{} changed from $40$ms to $120$ms.
Using the tuned $\gamma_{\delta t, 2} \approx 0.9227$ provided a modest recovery, and the additional scaling based on \dt{} regained a performance almost similar to the peak at ${\delta t}_0=40$ms.
The results indicate that tuning $\gamma$ for an initial ${\delta t}_0$ is essential for adapting $\gamma$ to different \dt{}s.
The high performance of $\delta t=120$ms with an atypical $\gamma_{\delta t, 1} \approx 0.9227^{120/40} = 0.786$ demonstrates the necessity of having guidelines such as our \delt{}-aware \gdt{} to avoid the poor performance of the baseline $\gamma=0.99$ when changing the \dt{}, especially since we observed that the asymptotic performance of $\gamma=0.99$ at $\delta t=120$ms remains lower than the peak even after extended learning.
Our \delt{}-aware hyper-parameters can thus lessen the demand for costly hyper-parameter tuning in the real world.

\section{Conclusions}

We introduced three benchmark tasks modified from existing tasks for cycle-time study and demonstrated that PPO may fail to learn using the baseline hyper-parameter values when the \dt{} changes.
Baseline hyper-parameter values performed significantly worse than their tuned values for both PPO and SAC.
We proposed a replacement set of \delt{}-aware hyper-parameters that adapt to different \dt{}s, and empirically showed that they perform as well as or substantially better than the baseline ones without re-tuning.
As opposed to the baseline hyper-parameters, the \delt{}-aware ones did not fail to learn with any \dt{}.
Our \delt{}-aware hyper-parameters can transfer hyper-parameter values tuned to a particular \dt{} to different \dt{}s on the same task without re-tuning and can lessen the need for extensive hyper-parameter tuning when \dt{} changes for both PPO and SAC, which is time-consuming and costly on real-world robots.
We finally validated the \delt{}-aware hyper-parameters on simulated and real-world robotic tasks.
Our extensive experiments amounted to 75 real-time hours or 10 million time steps of real-world robot learning, along with 20,000 and 1000 simulation hours for PPO and SAC respectively.

\newcommand{\hangin}{\goodbreak\hangindent=.35cm \noindent}

\section*{REFERENCES}

\hangin
Akkaya, I., Andrychowicz, M., Chociej, M., Litwin, M., McGrew, B., Petron, A., Paino, A., Plappert, M., Powell, G., Ribas, R., Schneider, J., Tezak, N., Tworek, J., Welinder, P., Weng, L., Yuan, Q., Zaremba, W., Zhang, L. (2019). Solving rubik's cube with a robot hand. \textit{arXiv preprint arXiv:1910.07113}.

\hangin
Baird, L. C. (1994). Reinforcement learning in continuous time: Advantage updating. In \textit{Proceedings of 1994 IEEE International Conference on Neural Networks}.

\hangin
Brockman, G., Cheung, V., Pettersson, L., Schneider, J., Schulman, J., Tang, J., Zaremba, W. (2016). Openai gym. \textit{arXiv preprint arXiv:1606.01540}.

\hangin
Chen, B., Xu, M., Li, L., Zhao, D. (2021). Delay-aware model-based reinforcement learning for continuous control. \textit{Neurocomputing 450}:119--128.

\hangin
Coumans, E., Bai, Y. (2016). PyBullet, a python module for physics simulation for games, robotics and machine learning. URL \url{http://pybullet.org}

\hangin
Doya, K. (2000). Reinforcement learning in continuous time and space. \textit{Neural Computation 12} (1):219--245.

\hangin
Dulac-Arnold, G., Levine, N., Mankowitz, D. J., Li, J., Paduraru, C., Gowal, S., Hester, T. (2021). Challenges of real-world reinforcement learning: Definitions, benchmarks and analysis. \textit{Machine Learning}. Advance online publication. \url{https://doi.org/10.1007/s10994-021-05961-4}.

\hangin
Farrahi, H. (2021). Investigating two policy gradient methods under different time discretizations. Master's thesis, Department of Computing Science, University of Alberta, Edmonton, AB T6G 2E8.

\hangin
Firoiu, V., Ju, T., Tenenbaum, J. (2018). At human speed: Deep reinforcement learning with action delay. \textit{arXiv preprint arXiv:\seqsplit{1810.07286}}.

\hangin
Fujimoto, S., Hoof, H., Meger, D. (2018). Addressing function approximation error in actor-critic methods. In \textit{International Conference on Machine Learning}.

\hangin
Gupta, A., Yu, J., Zhao, T. Z., Kumar, V., Rovinsky, A., Xu, K., Devlin, T., Levine, S. (2021). Reset-free reinforcement learning via multi-task learning: Learning dexterous manipulation behaviors without human intervention. In \textit{2021 IEEE International Conference on Robotics and Automation}.

\hangin
Haarnoja, T., Zhou, A., Hartikainen, K., Tucker, G., Ha, S., Tan, J., Kumar, V., Zhu, H., Gupta, A., Abbeel, P., Levine, S. (2018a). Soft actor-critic algorithms and applications. \textit{arXiv preprint arXiv:\seqsplit{1812.05905}}.

\hangin
Haarnoja, T., Ha, S., Zhou, A., Tan, J., Tucker, G., Levine, S. (2018b). Learning to walk via deep reinforcement learning. In \textit{Robotics: Science and Systems XV}.

\hangin
Keng, W. L., Graesser, L., Cvitkovic, M. (2019). SLM lab: A comprehensive benchmark and modular software framework for reproducible deep reinforcement learning. \textit{arXiv preprint arXiv:\seqsplit{1912.12482}}.

\hangin
Kingma, D. P., Ba, J. (2014). Adam: A method for stochastic optimization. In \textit{3rd International Conference on Learning Representations}.

\hangin
Lee, J., Sutton, R. S. (2021). Policy iterations for reinforcement learning problems in continuous time and space---fundamental theory and methods. \textit{Automatica 126}, Article 109421.

\hangin
Mahmood, A. R., Korenkevych, D., Vasan, G., Ma, W., Bergstra, J. (2018). Benchmarking reinforcement learning algorithms on real-world robots. In \textit{Proceedings of the 2nd Annual Conference on Robot Learning}.

\hangin
Munos, R. (2006). Policy gradient in continuous time. \textit{Journal of Machine Learning Research 7} (27):771--791.

\hangin
Raffin, A., Stulp, F. (2020). Generalized state-dependent exploration for deep reinforcement learning in robotics. \textit{arXiv preprint arXiv:\seqsplit{2005.05719}}.

\hangin
Ramstedt, S., Pal, C. (2019). Real-time reinforcement learning. In \textit{Advances in Neural Information Processing Systems}.

\hangin
Schulman, J., Wolski, F., Dhariwal, P., Radford, A., Klimov, O. (2017). Proximal policy optimization algorithms. \textit{arXiv preprint arXiv:\seqsplit{1707.06347}}.

\hangin
Sutton, R. S., Barto, A. G. (2018). \textit{Reinforcement Learning: An Introduction}. MIT Press.

\hangin
Tallec, C., Blier, L., Ollivier, Y. (2019). Making deep Q-learning methods robust to time discretization. In \textit{Proceedings of the 36th International Conference on Machine Learning}.

\hangin
Tan, J., Zhang, T., Coumans, E., Iscen, A., Bai, Y., Hafner, D., Bohez, S., Vanhoucke, V. (2018). Sim-to-real: Learning agile locomotion for quadruped robots. \textit{arXiv preprint arXiv:1804.10332}.

\hangin
Tassa, Y., Tunyasuvunakool, S., Muldal, A., Doron, Y., Trochim, P., Liu, S., Bohez, S., Merel, J., Erez, T., Lillicrap, T., Heess, N. (2020). dm\_control: Software and tasks for continuous control. \textit{arXiv preprint arXiv:2006.12983}.

\hangin
Travnik, J. B., Mathewson, K. W., Sutton, R. S., Pilarski, P. M. (2018). Reactive reinforcement learning in asynchronous environments. \textit{Frontiers in Robotics and AI 5}:79.

\clearpage
\appendix

\section*{Algorithms}

\subsection{The Standard Agent-Environment Interaction Loop}\label{app:standard_loop}
    \vspace{0pt}
\begin{algorithm}[h]

\DontPrintSemicolon
\SetAlgoNoLine

  $\Psi \doteq \mathrm{Initialize} ()$

  Retrieve from $\Psi$: learning period $U$, batch size $b$, parameterized policy $\pi_{\bm\theta}(a|s)$

  \BlankLine

  Initialize Buffer $B$ with capacity $b$
  
  Initialize $S_0 \sim d_0(\cdot)$

  \For{environment step $i=0, 1, 2, ...$} {
      Calculate action $A_i \sim \pi_{\bm\theta}(\cdot | S_i)$

      Apply $A_i$ and observe $R_{i+1}, S'_{i+1}$

      $T_{i+1} \doteq \mathbbm{1}_{S'_{i+1} \mathrm{\;is\;terminal}}$

      Store transaction in buffer $B_i = \left( S_i, A_i, R_{i+1}, S'_{i+1}, T_{i+1} \right)$

      \uIf{$i+1 \bmod U = 0$} {
          $\Psi \doteq \mathrm{Learn} (B, \Psi)$
      }

      \uIf{$T_{i+1}=1$} {
          Sample $S_{i+1} \sim d_0(\cdot)$
      }
      \uElse{
          $S_{i+1} \doteq S'_{i+1}$
      }

  }

  \caption{Agent-environment interaction loop}
  \label{alg:experience_collection}
\end{algorithm}
\hspace{.03\textwidth}
    \vspace{0pt}

\subsection{One-Step Actor-Critic}\label{app:one-step-ac}
\begin{algorithm}[h]

\DontPrintSemicolon
\SetAlgoNoLine

\SetKwFunction{FInitialize}{Initialize}
\SetKwFunction{FLearn}{Learn}
\SetKwProg{Fn}{Function}{:}{}

\Fn{\FInitialize{}} {

    $\Psi \doteq \{$ learning period $U$, batch size $b$, discount factor $\gamma$, parameterized policy $\pi_{\bm\theta}(a|s)$, parameterized value estimate $\hat{\upsilon}_\mathbf{w}(s)$, current discount $I$, policy learning rate $\eta^{\bm\theta}$, value estimate learning rate $\eta^\mathbf{w}$$\}$

    \BlankLine
    
    Denote network parameters by $\Psi.\bm\theta$, and $\Psi.\mathbf{w}$

    Initialize parameters $\Psi.\bm\theta$, $\Psi.\mathbf{w}$

    $I \doteq 1$

    \Return{$\Psi$}

}

\BlankLine
\BlankLine

\Fn{\FLearn{$B$, $\Psi$}} {

    Retrieve the only transaction in the buffer $S_t, A_t, R_{t+1}, S'_{t+1}, T_{t+1}$

    $\bm\theta \doteq \Psi.\bm\theta$ ; $\mathbf{w} \doteq \Psi.\mathbf{w}$
    
    $\overline{\mathbf{w}} \doteq \mathbf{w}$

    $\delta_t \doteq R_{t+1} + (1-T_{t+1}) \gamma \hat{\upsilon}_{\overline{\mathbf{w}}}(S'_{t+1}) - \hat{\upsilon}_{\mathbf{w}}(S_t)$

    $\bm\theta \doteq \bm\theta + \eta^{\bm\theta} I \delta_t \nabla_{\bm\theta} \log \pi_{\bm\theta}(A_t|S_t)$

    $\mathbf{w} \doteq \mathbf{w} + \eta^\mathbf{w} I \delta_t \nabla_\mathbf{w} \hat{\upsilon}_{\mathbf{w}}(S_t)$

    \lIf{$T_{i+1}=1$}{
        $I \doteq 1$
    }
    \lElse{
        $I \doteq \gamma I$
    }
    
    $\Psi.\bm\theta \doteq \bm\theta$ ; $\Psi.\mathbf{w} \doteq \mathbf{w}$
    
    \Return{$\Psi$}
  }

  \caption{One-step actor-critic Initialize and Learn functions}
  \label{alg:one_step_actor_critic_learn_function}
\end{algorithm}

\subsection{Proximal Policy Optimization}\label{app:ppo_learn_function}
Algorithm \ref{alg:ppo_learn_function} shows the \textit{Initialize} and \textit{Learn} functions of PPO. Hyper-parameters $N=10$ and $\epsilon=0.2$ remained fixed across all experiments. Batch size and the learning period were always equal $b=U$. We used the Adam optimizer (Kingma \& Ba 2014) with the same learning rate of $0.0003$ for both policy and value estimate objectives.
The architecture of the agent comprises a neural network of two hidden layers, each with $64$ units and $\tanh$ activations, producing the mean $\mu$, and state-independent parameters for the $\sigma$ of a normal distribution $\mathcal{N}(\mu, \sigma^2)$ from which the actions are sampled. The value estimate is parameterized by another neural network configured similarly to that which produces $\mu$. The equation below shows the gradient of the policy objective that is used in the algorithm.
\begin{align}
    \nabla_{\bm\theta} L_{\bm\theta,t} =
    \begin{cases}
        -\frac{\nabla_{\bm\theta} \pi_{\bm\theta}(A_t|S_t)}{\pi_{{\bm\theta}_{old}}(A_t|S_t)} \tilde{h}_t, &\parbox{10cm}{if $\rho_t(\bm\theta) \tilde{h}_t \leq \rho_t^{\mathrm{clip}}(\bm\theta) \tilde{h}_t$ \\ or $\Big( \rho_t(\bm\theta) \tilde{h}_t > \rho_t^{\mathrm{clip}}(\bm\theta) \tilde{h}_t$ and \\ $1-\epsilon \leq \rho_t(\bm\theta) \leq 1+\epsilon \Big)$} \\
        &\\
        0, &\text{otherwise}
    \end{cases} \label{eq:ppo_grad_L_theta}
\end{align}

\begin{algorithm}[h]

\DontPrintSemicolon
\SetAlgoNoLine

\SetKwFunction{FInitialize}{Initialize}
\SetKwFunction{FLearn}{Learn}
\SetKwProg{Fn}{Function}{:}{}

\Fn{\FInitialize{}} {

    $\Psi \doteq \{$ learning period $U$, number of epochs $N$, batch size $b$, mini-batch size $m$, discount factor $\gamma$, trace-decay parameter $\lambda$, clipping parameter $\epsilon$, parameterized policy $\pi_{\bm\theta}(a|s)$, parameterized value estimate $\hat{\upsilon}_\mathbf{w}(s)$, learning rate $\eta$ $\}$

    \BlankLine

    Denote network parameters by $\Psi.\bm\theta$, and $\Psi.\mathbf{w}$

    Initialize parameters $\Psi.\bm\theta$, $\Psi.\mathbf{w}$

    \Return{$\Psi$}

}

\BlankLine
\BlankLine

\Fn{\FLearn{$B$, $\Psi$}} {

    $\bm\theta \doteq \Psi.\bm\theta$ ; $\mathbf{w} \doteq \Psi.\mathbf{w}$

    $\overline{\mathbf{w}} \doteq \mathbf{w}$

    \For{$t=0, 1, 2, ..., b$} {
  
      Retrieve transaction from buffer $S_t, A_t, R_{t+1}, S'_{t+1}, T_{t+1}$
  
      $T \doteq \min \left\{ j \mid j \in \mathbb{N}\ \land\ j > t\ \land\ T_j=1 \right\}$
      
      $G_t^\lambda \doteq \lambda^{T-t-1} G_t + (1-\lambda) \sum_{n=1}^{T-t-1} \lambda^{n-1} G_{t:t+n}$,

      \quad where $G_{t:t+n} \doteq \gamma^n \hat{\upsilon}_{\overline{\mathbf{w}}}(S_{t+n}) + \sum_{j=1}^n \gamma^{j-1} R_{t+j}$
      
      $\hat{h}_t \doteq G_t^\lambda - \hat{\upsilon}_\mathbf{w}(S_t)$ \tcp*[l]{advantage estimate}
    }
    
    $\tilde{h} \doteq \textrm{normalize}(\hat{h})$
    
    $D \doteq \left( (S_t, A_t, \tilde{h}_t, G_t^\lambda) \right)_{t=0}^{b}$
    
    $\bm\theta_{old} \doteq \bm\theta$
    
    \For{epoch $e=1, 2, ..., N$} {
      $\tilde{D} \doteq \textrm{shuffle}(D)$
      
      Slice $\tilde{D}$ into $\lceil \frac{b}{m} \rceil$ mini-batches
      
      \For{each mini-batch $M$} {

        $\bm\theta \doteq \bm\theta - \eta \frac{1}{m} \sum_{\left(S_k,A_k,\tilde{h}_k\right) \in M} \nabla_{\bm\theta} L_{\bm\theta,k}$

        \quad using (\ref{eq:ppo_grad_L_theta}), where

        \quad $L_{\bm\theta,k} \doteq - \min\left( \rho_k(\bm\theta) \tilde{h}_k, \rho_k^{\mathrm{clip}}(\bm\theta) \tilde{h}_k \right)$,

        \quad $\rho_k(\bm\theta) \doteq \frac{\pi_{\bm\theta}(A_k|S_k)}{\pi_{{\bm\theta}_{old}}(A_k|S_k)}$, and

        \quad $\rho_k^{\mathrm{clip}}(\bm\theta) \doteq \mathrm{clip}(\rho_k(\bm\theta), 1-\epsilon, 1+\epsilon)$

        $\mathbf{w} \doteq \mathbf{w} + \eta \frac{1}{m} \sum_{\left(S_k,G_k^\lambda\right) \in M} 2 \left( {G_k^{\lambda}} - \hat{\upsilon}_\mathbf{w}(S_k) \right) \nabla_\mathbf{w} \hat{\upsilon}_\mathbf{w}(S_k)$
      }
    }

    $\Psi.\bm\theta \doteq \bm\theta$ ; $\Psi.\mathbf{w} \doteq \mathbf{w}$

    \Return{$\Psi$}
  }

  \caption{PPO Initialize and Learn functions}
  \label{alg:ppo_learn_function}
\end{algorithm}

\clearpage
\subsection{Soft Actor-Critic}\label{app:sac_learn_function}

Algorithm \ref{alg:sac_learn_function} shows the \textit{Initialize} and \textit{Learn} functions of SAC. Hyper-parameters $b=1,\!000,\!000$, $U=1$, $m=256$, $g=1$, and $\tau=0.005$ remained constant for all experiments. We used the Adam optimizer (Kingma \& Ba 2014) for all objectives with learning rates set to $0.0003$.
Both the policy and the action-value estimates were parameterized using neural networks with two hidden layers of size $256$ with ReLU activations. The hidden layers of the policy parameters were shared between mean $\mu$ and standard deviation $\sigma$ of the normal distribution $\mathcal{N}(\mu, \sigma^2)$ that was used to draw the actions.
Following equations are the gardients of all objectives that are used in the algorithm.

\begin{align}
    \nabla_{\bm\theta} L_{\bm\theta,t} =\, &\alpha \nabla_{\bm\theta} \log \pi_{\bm\theta}(a|S_t) \vert_{a=f(S_t,\epsilon_t,\bm\theta)} \nonumber \\
    &+ \nabla_{\bm\theta} f(S_t,\epsilon_t,\bm\theta) \Big( \alpha \nabla_a \log \pi_{\bm\theta}(a|S_t) \vert_{a=f(S_t,\epsilon_t,\bm\theta)} \nonumber \\
    &- \nabla_a \hat{q}_{\mathbf{w}_\mathrm{min}}(S_t,a) \vert_{a=f(S_t,\epsilon_t,\bm\theta)} \Big), \text{ where} \label{eq:sac_grad_L_theta} \\
    \mathbf{w}_\mathrm{min} \doteq &\argmin_{\mathbf{w} \in \{\mathbf{w}_1, \mathbf{w}_2\}} \hat{q}_\mathbf{w}(S_t,f(S_t,\epsilon_t,\bm\theta)), \nonumber
\end{align}

\begin{align}
    \nabla_{\mathbf{w}_i} L_{\mathbf{w}_i,t} = \Big( \hat{q}_{\mathbf{w}_i} (S_t, A_t) - &\big( R_{t+1} + (1-T_{t+1}) \gamma V(S'_{t+1}) \big) \Big) \nonumber \\ &\nabla_{\mathbf{w}_i} \hat{q}_{\mathbf{w}_i}(S_t, A_t) \text{ for } i \in \{1,2\}, \label{eq:sac_grad_L_wi}
\end{align}
\begin{equation}
    \nabla_\alpha L_{\alpha,t} = - \log \pi_{\bm\theta}(f(S_t, \epsilon_t, \bm\theta) | S_t) - \overline{\mathcal{H}}. \label{eq:sac_grad_L_alpha}
\end{equation}

\begin{algorithm}[h]

\DontPrintSemicolon
\SetAlgoNoLine

\SetKwFunction{FInitialize}{Initialize}
\SetKwFunction{FLearn}{Learn}
\SetKwProg{Fn}{Function}{:}{}

\Fn{\FInitialize{}} {

    $\Psi \doteq \{$ learning period $U$, batch size $b$, mini-batch size $m$, gradient steps $g$, discount factor $\gamma$, temperature parameter $\alpha$, target smoothing coefficient $\tau$, parameterized policy $\pi_{\bm\theta}(a|s)$, policy mean $\mu_{\bm\theta}(s)$, policy standard deviation $\sigma_{\bm\theta}(s)$, parameterized action-value estimates $\hat{q}_{\mathbf{w}_1}(s,a)$, $\hat{q}_{\mathbf{w}_2}(s,a)$, $\hat{q}_{\overline{\mathbf{w}}_1}(s,a)$, $\hat{q}_{\overline{\mathbf{w}}_2}(s,a)$, policy learning rate $\eta^{\bm\theta}$, action-value estimate learning rate $\eta^\mathbf{w}$, temperature learning rate $\eta^\alpha$ $\}$

    \BlankLine

    Denote network parameters by $\Psi.\bm\theta$, $\Psi.\mathbf{w}_1$, $\Psi.\mathbf{w}_2$, $\Psi.\overline{\mathbf{w}}_1$, $\Psi.\overline{\mathbf{w}}_2$, and $\Psi.\alpha$

    Initialize parameters $\Psi.\bm\theta$, $\Psi.\mathbf{w}_1$, and $\Psi.\mathbf{w}_2$

    $\Psi.\overline{\mathbf{w}}_1 \doteq \Psi.\mathbf{w}_1$ ; $\Psi.\overline{\mathbf{w}}_2 \doteq \Psi.\mathbf{w}_2$ ; $\Psi.\alpha \doteq 1$

    \Return{$\Psi$}
}

\BlankLine
\BlankLine

\Fn{\FLearn{$B$, $\Psi$}} {

    $\left( \bm\theta, \mathbf{w}_1, \mathbf{w}_2, \overline{\mathbf{w}}_1, \overline{\mathbf{w}}_2, \alpha \right) \doteq \Psi.\left( \bm\theta, \mathbf{w}_1, \mathbf{w}_2, \overline{\mathbf{w}}_1, \overline{\mathbf{w}}_2, \alpha \right)$

    \For{each gradient step from $1$ to $g$} {

        Sample a mini-batch $M$ of size $m$ uniformly randomly from $B$

        $\mathbf{w}_1 \doteq \mathbf{w}_1 - \eta^\mathbf{w} \frac{1}{m} \sum_{\left(S_k,A_k,R_{k+1},S'_{k+1},T_{k+1}\right) \in M} \nabla_{\mathbf{w}_1} L_{\mathbf{w}_1,k}$

        $\mathbf{w}_2 \doteq \mathbf{w}_2 - \eta^\mathbf{w} \frac{1}{m} \sum_{\left(S_k,A_k,R_{k+1},S'_{k+1},T_{k+1}\right) \in M} \nabla_{\mathbf{w}_2} L_{\mathbf{w}_2,k}$

        \quad using (\ref{eq:sac_grad_L_wi}), where

        \quad $L_{\mathbf{w}_1,k} \doteq \frac{1}{2} \Big( \hat{q}_{\mathbf{w}_1}(S_k, A_k) - \big( R_{k+1} + (1-T_{k+1}) \gamma V(S'_{k+1}) \big) \Big)^2$,

        \quad $L_{\mathbf{w}_2,k} \doteq \frac{1}{2} \Big( \hat{q}_{\mathbf{w}_2}(S_k, A_k) - \big( R_{k+1} + (1-T_{k+1}) \gamma V(S'_{k+1}) \big) \Big)^2$,

        \quad $\begin{aligned}
            V(S'_{k+1}) \doteq &\min\left( \hat{q}_{\overline{\mathbf{w}}_1} (S'_{k+1}, \tilde{A}_{k+1}), \hat{q}_{\overline{\mathbf{w}}_2} (S'_{k+1}, \tilde{A}_{k+1}) \right) \\
                               &- \alpha \log \pi_{\bm\theta}(\tilde{A}_{k+1}|S'_{k+1}),
        \end{aligned}$

        \quad and $\tilde{A}_{k+1} \sim \pi_{\bm\theta}(\cdot | S'_{k+1}) \doteq \mathcal{N}\left(\mu_{\bm\theta}(S'_{k+1}), {\sigma_{\bm\theta}(S'_{k+1})}^2\right)$

        $\bm\theta \doteq \bm\theta - \eta^{\bm\theta} \frac{1}{m} \sum_{\left(S_k,A_k,R_{k+1},S'_{k+1},T_{k+1}\right) \in M} \nabla_{\bm\theta} L_{\bm\theta,k}$

        \quad using (\ref{eq:sac_grad_L_theta}), where

        \quad $\begin{aligned}
            L_{\bm\theta,k} &\doteq \alpha \log \pi_{\bm\theta}(f(S_k, \epsilon_k, \bm\theta) | S_k) \\
                                   &- \min \Big( \hat{q}_{\mathbf{w}_1} (S_k, f(S_k, \epsilon_k, \bm\theta)), \hat{q}_{\mathbf{w}_2} (S_k, f(S_k, \epsilon_k, \bm\theta)) \Big),
        \end{aligned}$

        \quad $f(S_k, \epsilon_k, \bm\theta) \doteq \mu_{\bm\theta}(S_k) + \epsilon_k\, \sigma_{\bm\theta}(S_k)$, \tcp*[l]{reparameterization}

        \quad and $\epsilon_k \sim \mathcal{N}(0, 1)$

        $\alpha \doteq \alpha - \eta^\alpha \frac{1}{m} \sum_{\left(S_k,A_k,R_{k+1},S'_{k+1},T_{k+1}\right) \in M} \nabla_\alpha L_{\alpha,k}$

        \quad using (\ref{eq:sac_grad_L_alpha}), where

        \quad $L_{\alpha,k} \doteq -\alpha \log \pi_{\bm\theta}(f(S_k, \epsilon_k, \bm\theta) | S_k) - \alpha \overline{\mathcal{H}}$,

        \quad and $\overline{\mathcal{H}} \doteq -|\mathcal{A}|$

        $\overline{\mathbf{w}}_1 \doteq \tau \mathbf{w}_1 + (1-\tau)\overline{\mathbf{w}}_1$

        $\overline{\mathbf{w}}_2 \doteq \tau \mathbf{w}_2 + (1-\tau)\overline{\mathbf{w}}_2$

    }

    $\Psi.\left( \bm\theta, \mathbf{w}_1, \mathbf{w}_2, \overline{\mathbf{w}}_1, \overline{\mathbf{w}}_2, \alpha \right) \doteq \left( \bm\theta, \mathbf{w}_1, \mathbf{w}_2, \overline{\mathbf{w}}_1, \overline{\mathbf{w}}_2, \alpha \right)$

    \Return{$\Psi$}
  }

  \caption{SAC Initialize and Learn functions}
  \label{alg:sac_learn_function}
\end{algorithm}

\clearpage

\section*{Hyper-Parameters}\label{app:hps}

\begin{table}[h]

\caption{Baseline Hyper-Parameters of PPO on the Reacher and the Double Pendulum Tasks.} \label{tbl:pposimhps}
\centering
\begin{tabular}{|c|c|}
\hline
baseline & \multirow{2}{*}{value} \\
hyper-parameters &  \\ \hline
$b$ & 2000  \\ \hline
$m$ & 50    \\ \hline
$\gamma$ & 0.99  \\ \hline
$\lambda$ & 0.95  \\ \hline
\end{tabular}

\bigskip
\caption{Tuned Hyper-Parameters of PPO on the Reacher Task.} \label{tbl:pposimtunedhps}
\centering
\begin{tabular}{|c|c|c|c|c|c|}
\hline
tuned & \multicolumn{5}{c|}{$\delta t$ (ms)} \\
hyper-parameters & 4 & 8 & 16  & 32  & 64  \\ \hline
$b$ & 8000 & 8000 & 4000 & 2000 & 1000  \\ \hline
$m$ & 50 & 25 & 12 & 12 & 12      \\ \hline
\end{tabular}

\bigskip
\caption{Baseline and $\delta t$-Aware Hyper-Parameters of PPO on Reacher and Double Pendulum Tasks.} \label{tbl:pposimhps_dtaware}
\centering
\begin{tabular}{|c|c|c|}
\hline
\multicolumn{2}{|c|}{hyper-parameters} & \multirow{2}{*}{value} \\
\cline{1-2}
baseline & \delt{}-aware &  \\ \hline
$b$ & $b_{16}$ & 2000  \\ \hline
$m$ & $m_{16}$ & 50    \\ \hline
$\gamma$ & $\gamma_{16}$ & 0.99  \\ \hline
$\lambda$ & $\lambda_{16}$ & 0.95  \\ \hline
\end{tabular}

\bigskip
\caption{Tuned $\gamma$ of SAC on the Reacher Task.} \label{tbl:sacsimtunedhps}
\centering
\begin{tabular}{|c|c|c|c|c|c|}
\hline
tuned & \multicolumn{5}{c|}{$\delta t$ (ms)} \\
hyper-parameter & 4 & 8 & 16  & 32  & 64  \\ \hline
$\gamma$ & 0.961 & 0.923 & 0.851 & 0.851 & 0.851  \\ \hline
\end{tabular}

\bigskip
\caption{Hyper-Parameters of PPO on the Real-Robot Reacher Task.} \label{tbl:ppour5hps}
\centering
\begin{tabular}{|c|c|c|}
\hline
\multicolumn{2}{|c|}{hyper-parameters} & \multirow{2}{*}{value} \\
\cline{1-2}
baseline & \delt{}-aware &  \\ \hline
$b$ & $b_{40}$ & 400  \\ \hline
$m$ & $m_{40}$ & 10    \\ \hline
$\gamma$ & $\gamma_{40}$ & 0.99  \\ \hline
$\lambda$ & $\lambda_{40}$ & 0.95  \\ \hline
\end{tabular}

\end{table}

\section*{Additional Figures}\label{app:learning_curves}

\begin{figure}[t]
    \centering
    \includegraphics[keepaspectratio=true, width=0.48\textwidth]{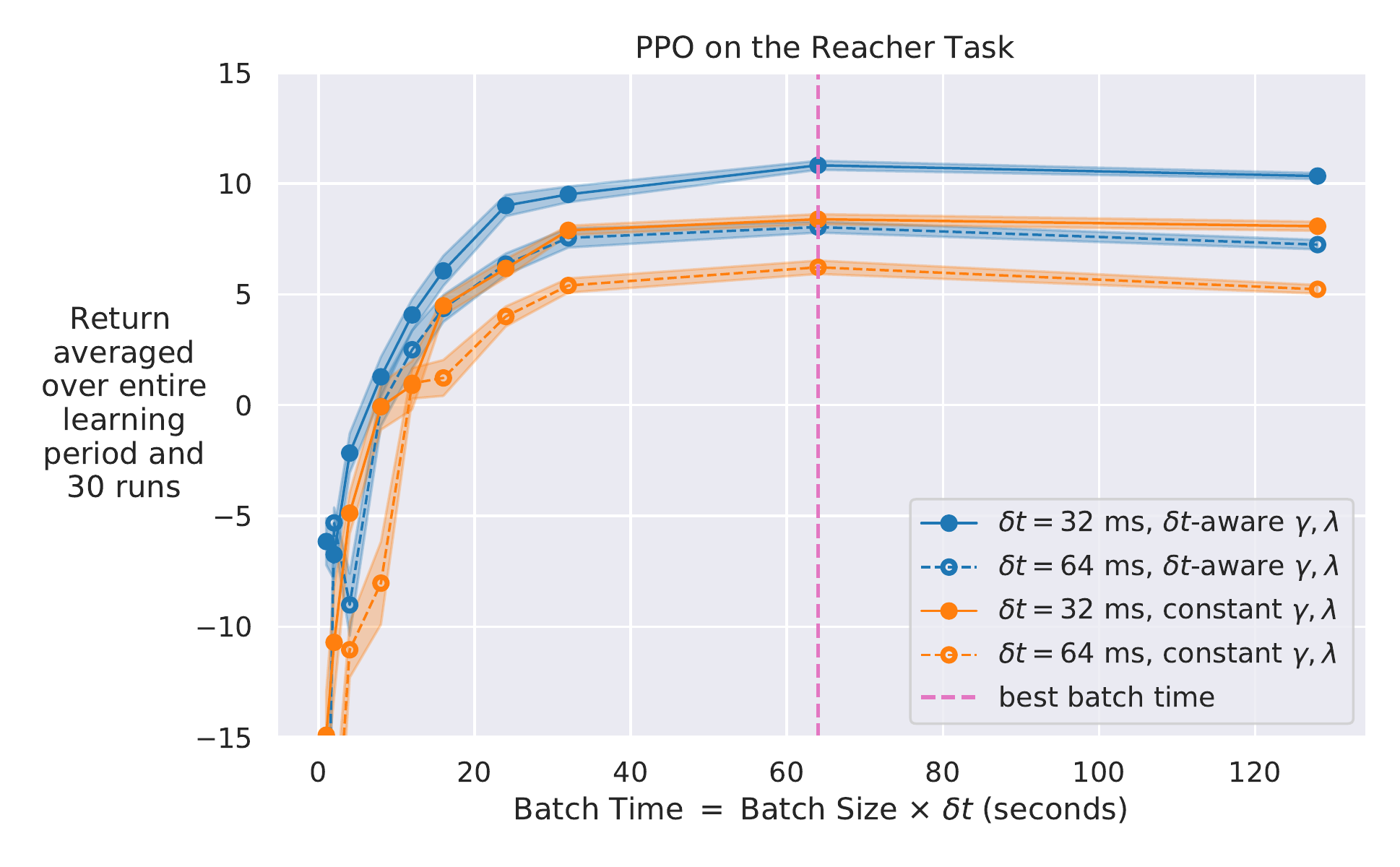}
        \caption[Overall average return of PPO comparing the \delt{}-aware hyper-parameters with constant $\gamma$ and $\lambda$] {Overall average return of \delt{}-aware hyper-parameters compared with ones where $\gamma_{\delta t} = \gamma_{16} = 0.99$ and $\lambda_{\delta t} = \lambda_{16} = 0.95$ are constant in all runs (constant).
        }
        \label{fig:auc_vs_bt_noscale_simple}
\end{figure}

\begin{figure*}[h]
    \centering
    \includegraphics[keepaspectratio=true, width=0.85\textwidth]{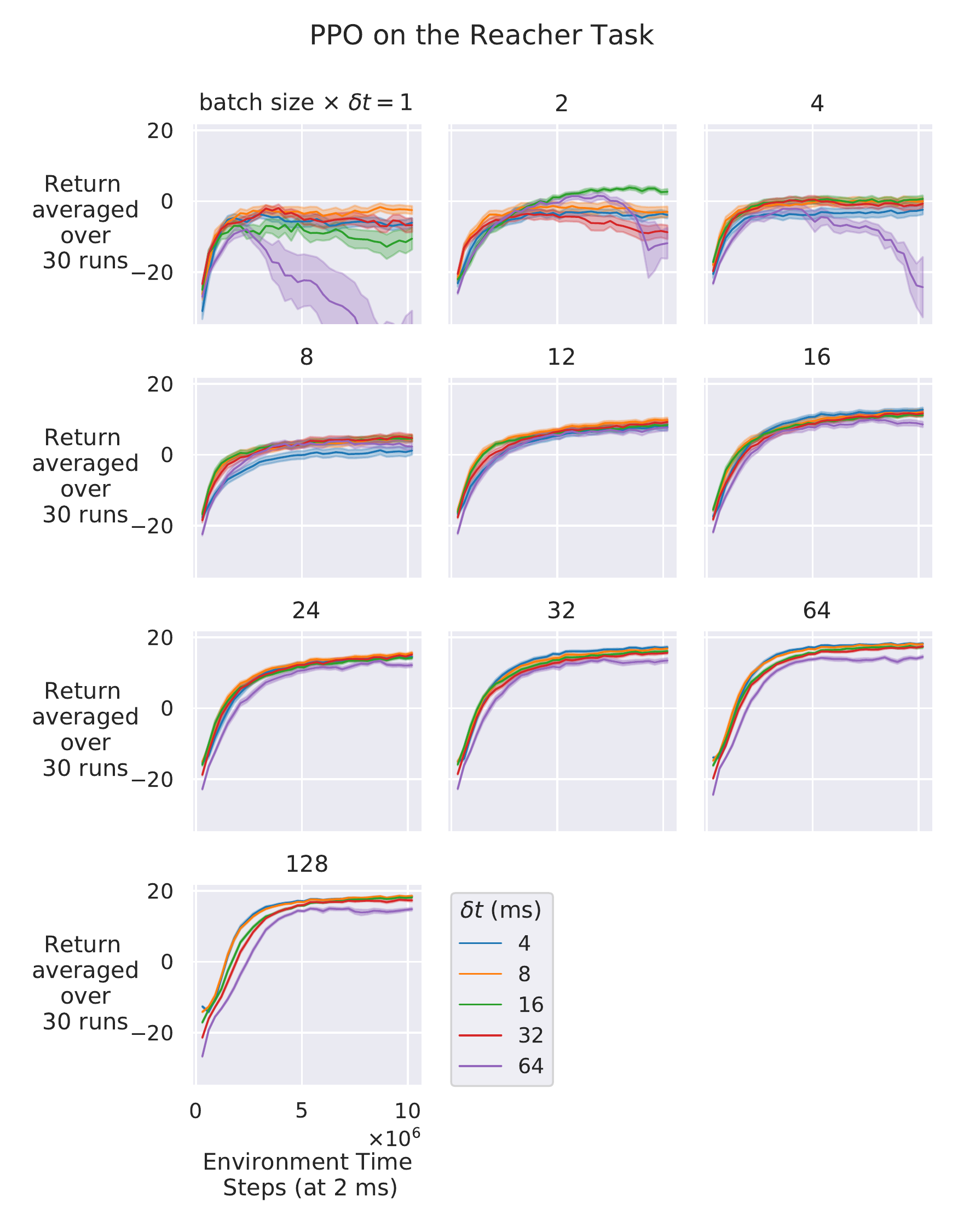}
    \caption[Learning curves of PPO using the \delt{}-aware hyper-parameters] {Corresponding learning curves for Figure \ref{fig:auc_vs_bt_final} using the \delt{}-aware hyper-parameters. Performance improves with increasing batch times for all \dt{}s.}
    \label{fig:auc_vs_bt_final_learning_curves}
\end{figure*}

\begin{figure*}
    \centering
    \includegraphics[keepaspectratio=true, width=0.85\textwidth]{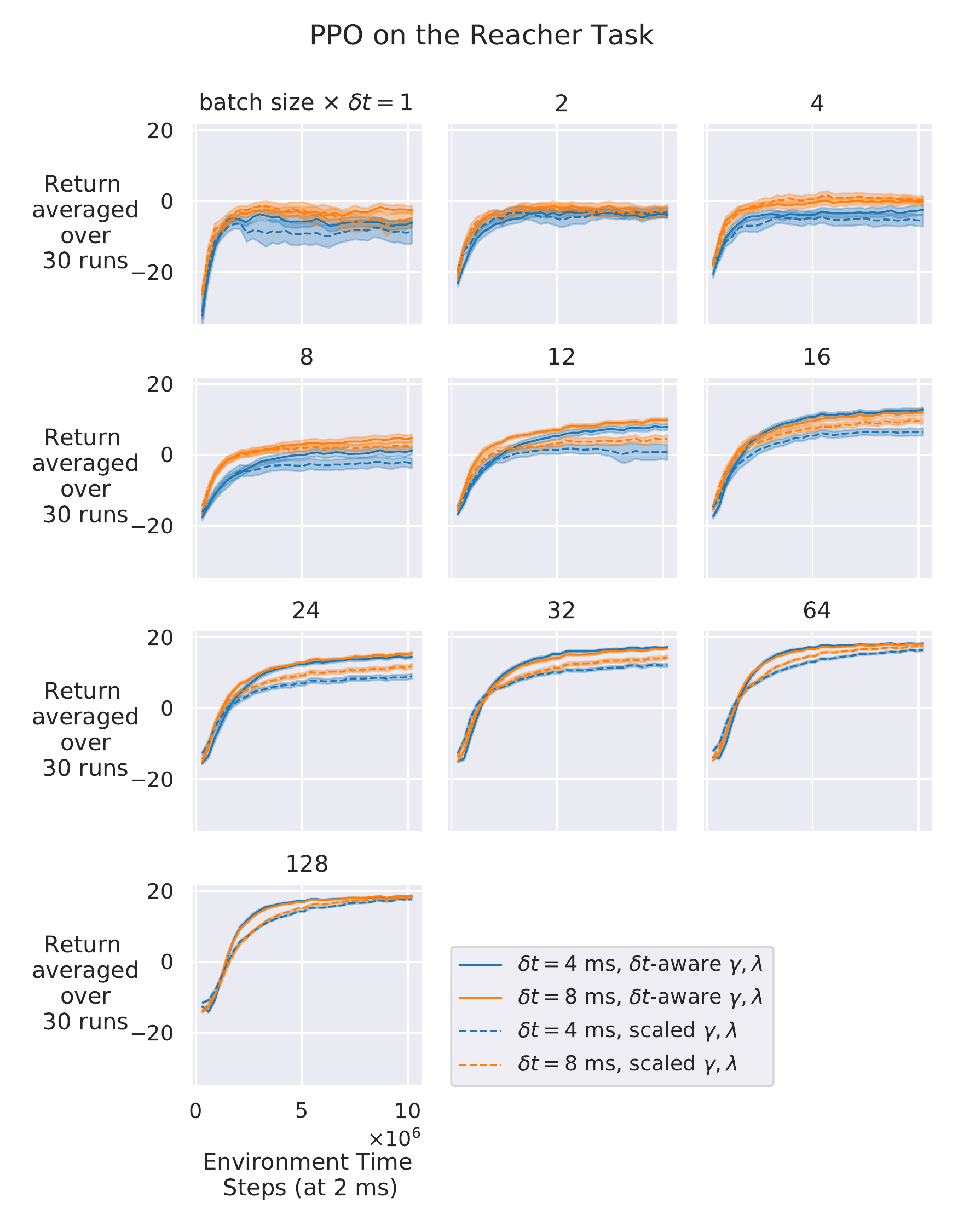}
    \caption[Learning curves of PPO comparing the \delt{}-aware hyper-parameters with ones where $\gamma$ and $\lambda$ are always exponentiated] {Corresponding learning curves for Figure \ref{fig:auc_vs_bt_simple} comparing the \delt{}-aware hyper-parameters with ones where the discount factor $\gamma$ and trace-decay parameter $\lambda$ are always exponentiated to the ${\delta t}/{{\delta t}_0}$ power (scaled). Only the $\delta t=4$ms and $\delta t=8$ms curves are plotted since the rest are similar to Figure \ref{fig:auc_vs_bt_final_learning_curves}.}
    \label{fig:auc_vs_bt_learning_curves}
\end{figure*}

\begin{figure*}
    \centering
    \includegraphics[keepaspectratio=true, width=0.85\textwidth]{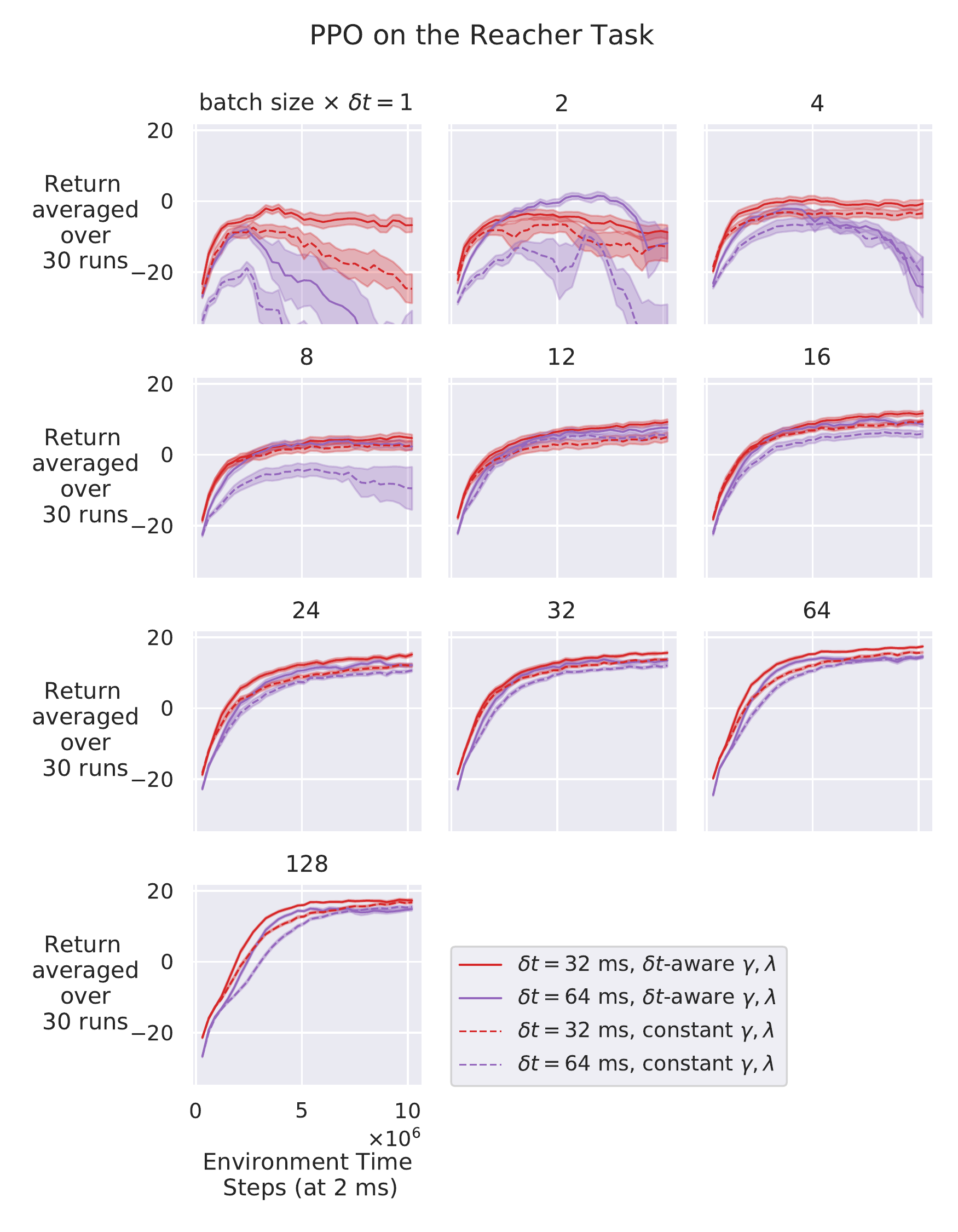}
    \caption[Learning curves of PPO comparing the \delt{}-aware hyper-parameters with ones where $\gamma$ and $\lambda$ are kept constant] {Corresponding learning curves for Figure \ref{fig:auc_vs_bt_noscale_simple} comparing the \delt{}-aware hyper-parameters with ones where the discount factor $\gamma_{\delta t} = \gamma_{16} = 0.99$ and trace-decay parameter $\lambda_{\delta t} = \lambda_{16} = 0.95$ are constant in all runs (constant). Only the $\delta t=32$ms and $\delta t=64$ms curves are plotted since the rest are similar to Figure \ref{fig:auc_vs_bt_final_learning_curves}.}
    \label{fig:auc_vs_bt_noscale_learning_curves}
\end{figure*}

\begin{figure*}
    \centering
    \includegraphics[keepaspectratio=true, width=0.85\textwidth]{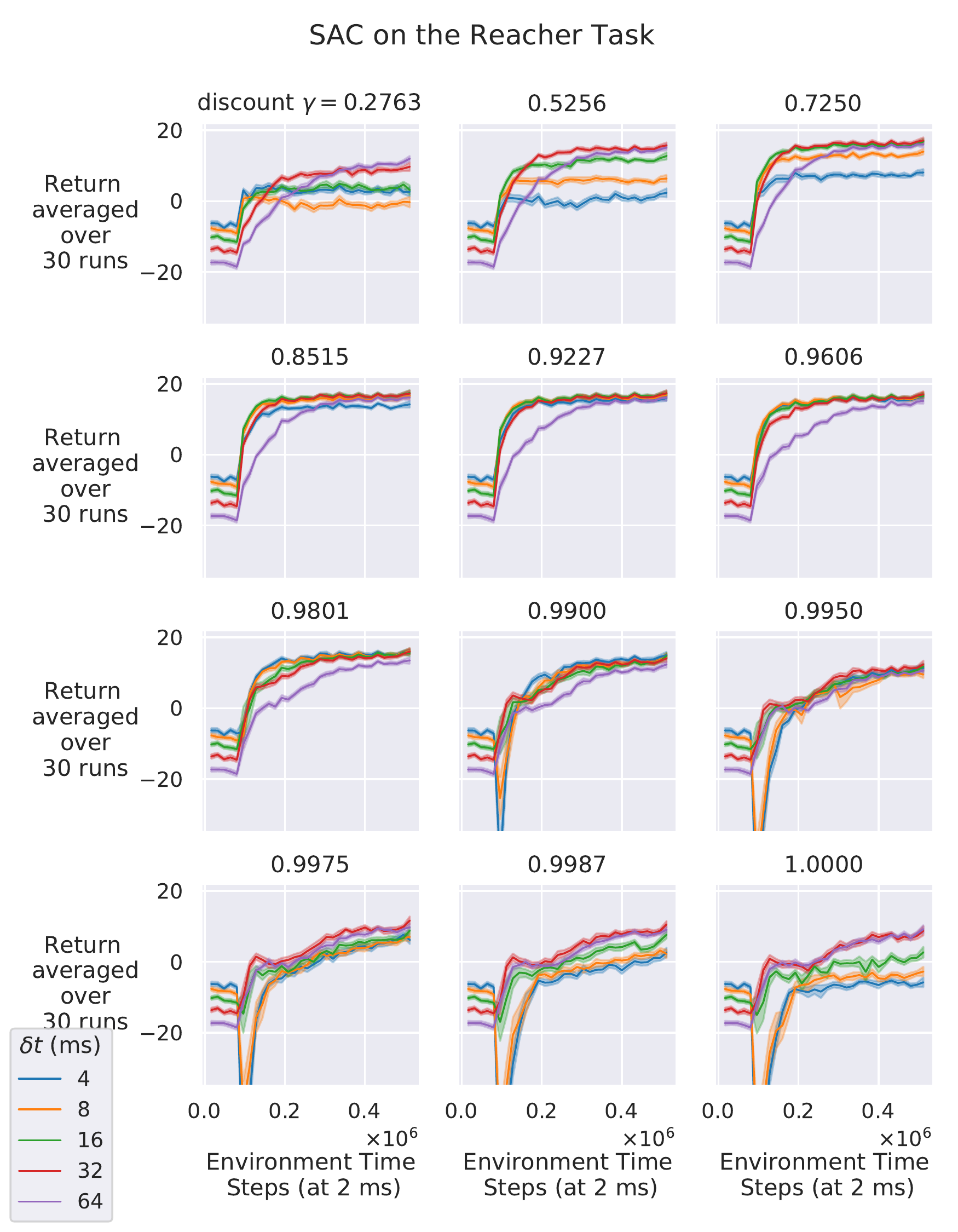}
    \caption[Learning curves of SAC for a sweep of $\gamma$ values] {Corresponding learning curves of SAC for Figure \ref{fig:sac_oar_vs_gam_256_reacher} for a sweep of $\gamma$ values. Intermediate values of $\gamma$ obtain better asymptotic performance and learning speed.}
    \label{fig:sac_oar_vs_gam_256_reacher_learning_curves}
\end{figure*}

\begin{figure*}[t]
    \centering
    \begin{minipage}[t]{.48\textwidth}
        \centering
        \includegraphics[keepaspectratio=true, width=1.0\textwidth]{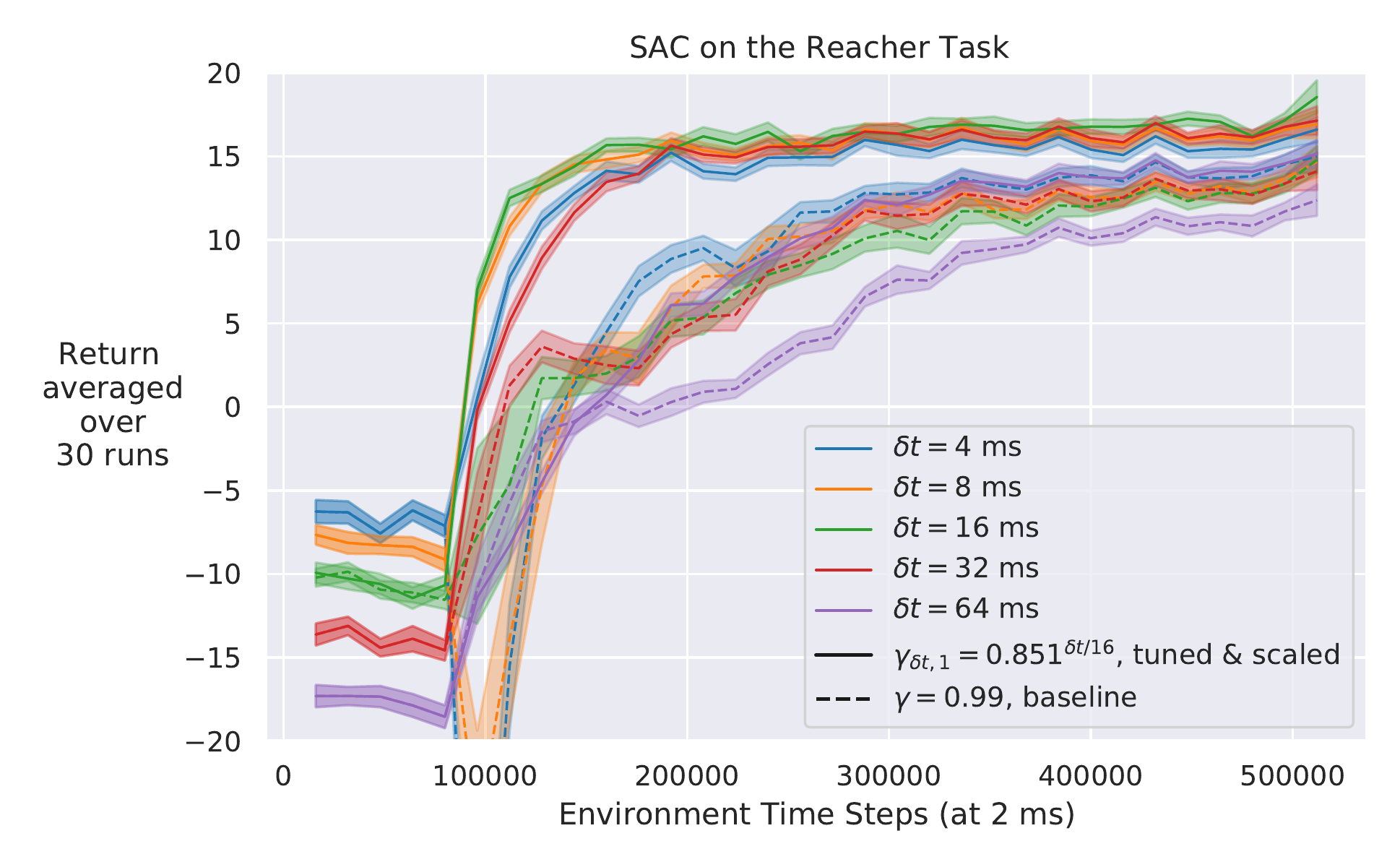}
        \caption[Learning curves of SAC comparing the baseline with the tuned and scaled $\gamma$] {Corresponding learning curves of SAC for Figure \ref{fig:icra2022_oar_vs_dt_combo} (left) comparing the baseline with the tuned and scaled $\gamma_{\delta t, 1}=0.851^{\delta t/16}$. All \dt{}s learn faster and avoid the sharp drops of smaller \dt{}s with the baseline $\gamma$.}
        \label{fig:sac_oar_vs_dt_hyps_reacher_tunescale_learning_curves}
    \end{minipage}\hfill
    \begin{minipage}[t]{.48\textwidth}
        \centering
        \includegraphics[keepaspectratio=true, width=1.0\textwidth]{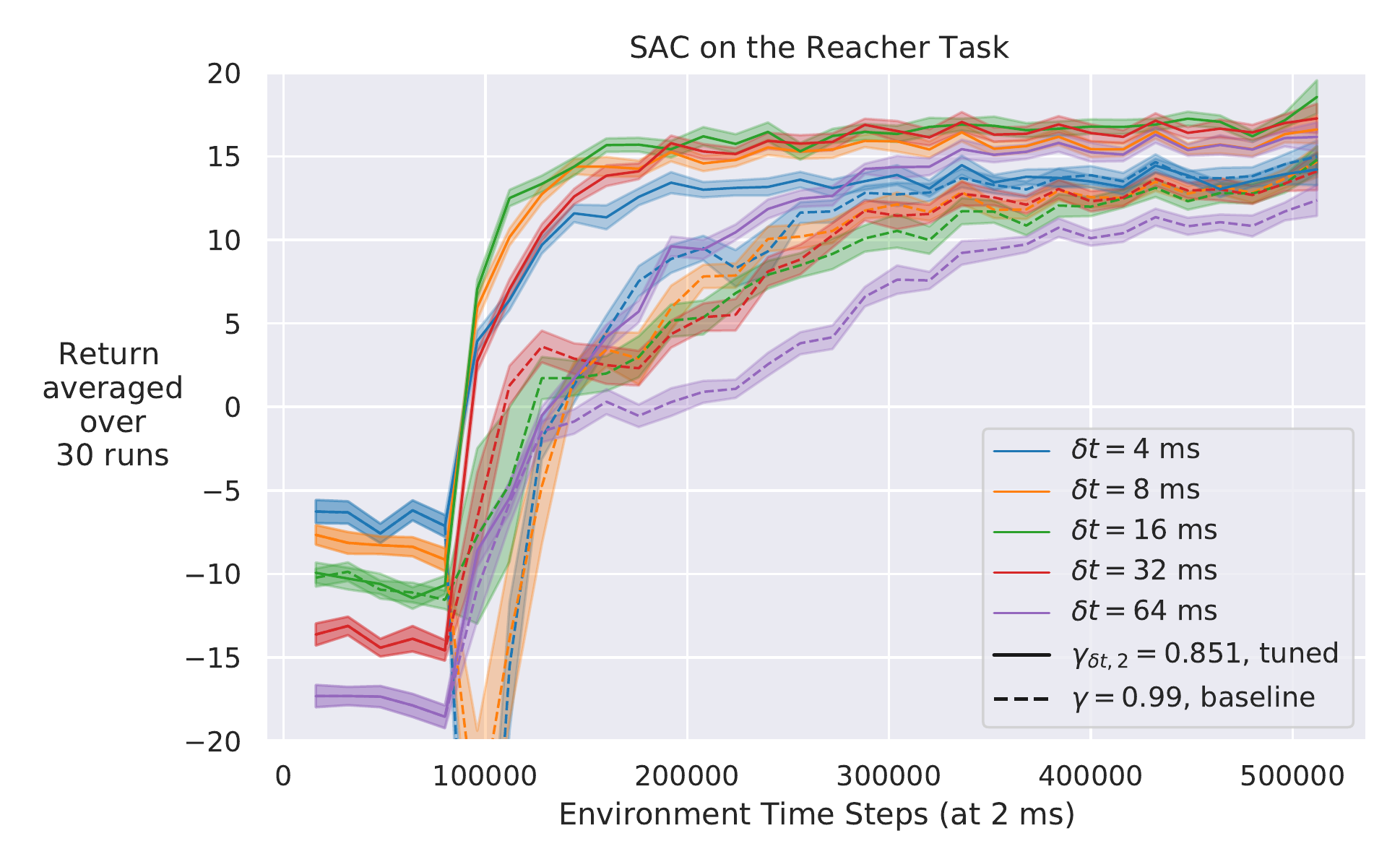}
        \caption[Learning curves of SAC comparing the baseline with the tuned $\gamma$] {Corresponding learning curves of SAC for Figure \ref{fig:icra2022_oar_vs_dt_combo} (left) comparing the baseline with the tuned $\gamma_{\delta t, 2}=0.851$. All \dt{}s learn faster and avoid the sharp drops of smaller \dt{}s with the baseline $\gamma$.}
        \label{fig:sac_oar_vs_dt_hyps_reacher_tune_learning_curves}
    \end{minipage}
\end{figure*}

\begin{figure*}[t]
    \centering
    \begin{minipage}[t]{.48\textwidth}
        \centering
        \includegraphics[keepaspectratio=true, width=1.0\textwidth]{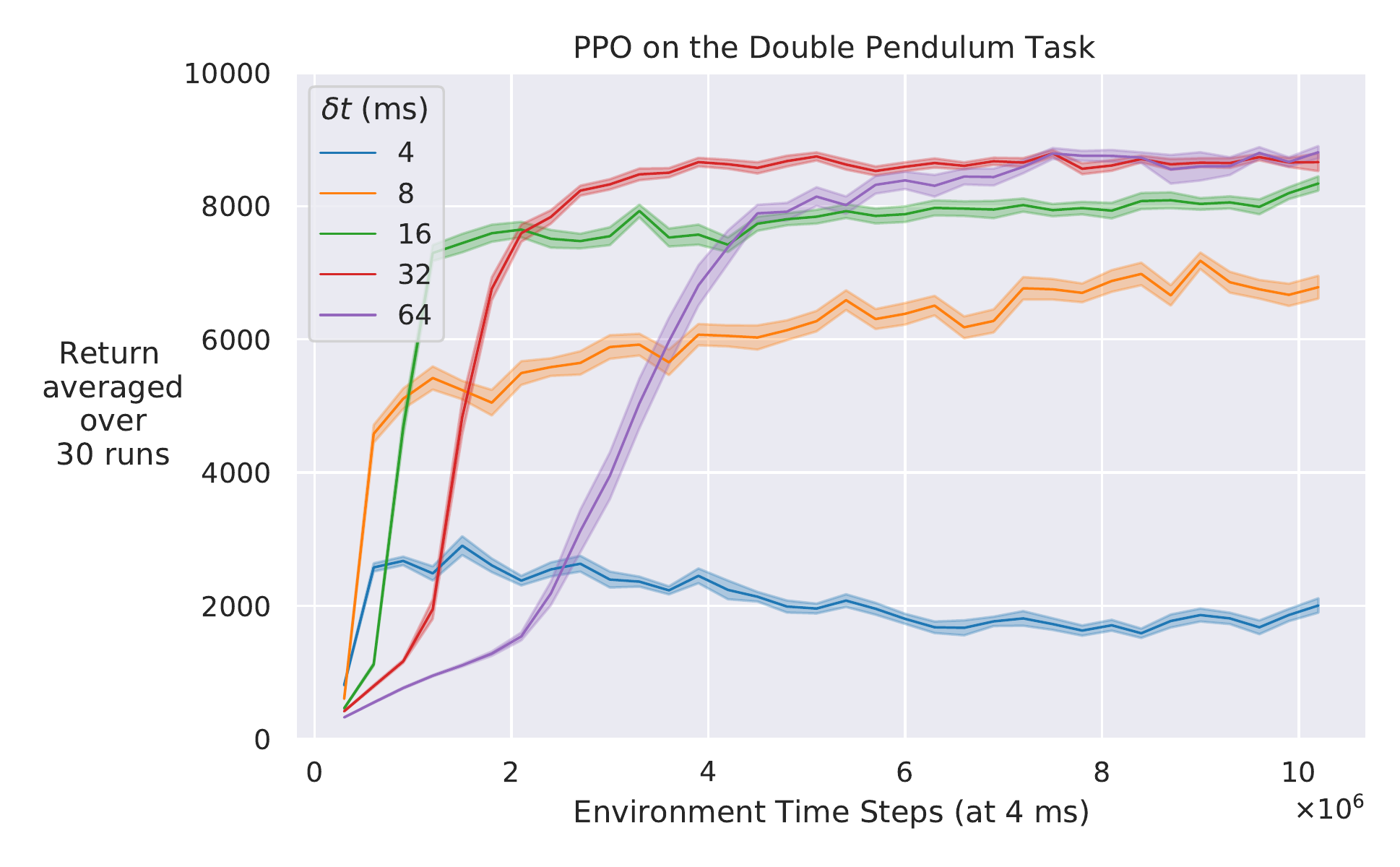}
        \caption[Learning curves of PPO using the baseline hyper-parameters on the Double Pendulum Task] {Corresponding learning curves of PPO for Figure \ref{fig:icra2022_oar_vs_dt_combo} (middle) using the baseline hyper-parameters on the Double Pendulum Task. Different \dt{}s vary greatly in asymptotic performance and learning speed.}
        \label{fig:invdblpndlm_defaults_vs_heuristics_auc_2000_default_learning_curves}
    \end{minipage}\hfill
    \begin{minipage}[t]{.48\textwidth}
        \centering
        \includegraphics[keepaspectratio=true, width=1.0\textwidth]{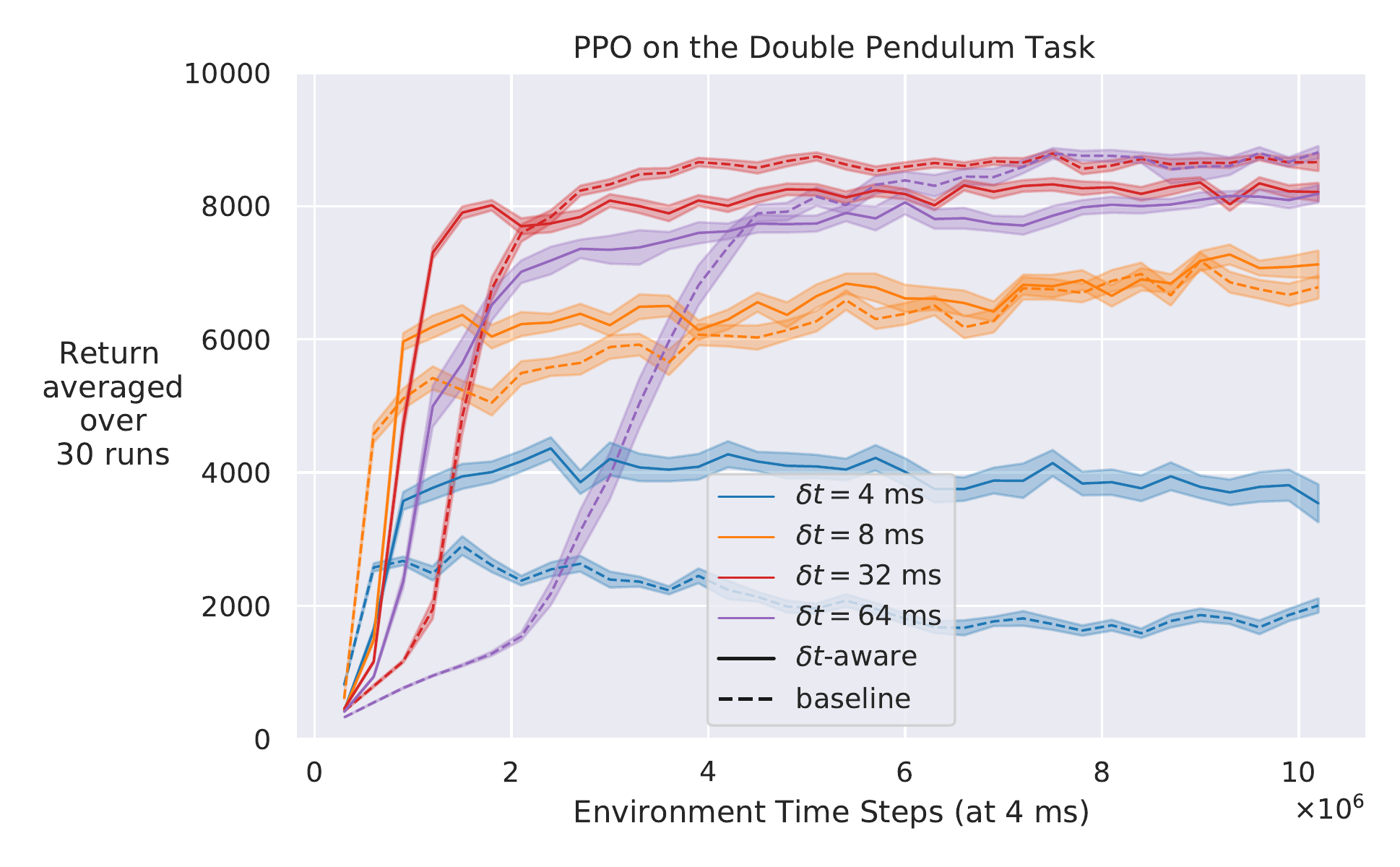}
        \caption[Learning curves of PPO comparing the baseline and \delt{}-aware hyper-parameters on the Double Pendulum Task] {Corresponding learning curves of PPO for Figure \ref{fig:icra2022_oar_vs_dt_combo} (middle) comparing the baseline and \delt{}-aware hyper-parameters on the Double Pendulum Task. With the \delt{}-aware hyper-parameters, different \dt{}s are more similar in asymptotic performance and learning speed.}
        \label{fig:invdblpndlm_defaults_vs_heuristics_auc_2000_heuristics_learning_curves}
    \end{minipage}
\end{figure*}

\begin{figure*}[t]
    \centering
    \begin{minipage}[t]{.48\textwidth}
        \centering
        \includegraphics[keepaspectratio=true, width=1.0\textwidth]{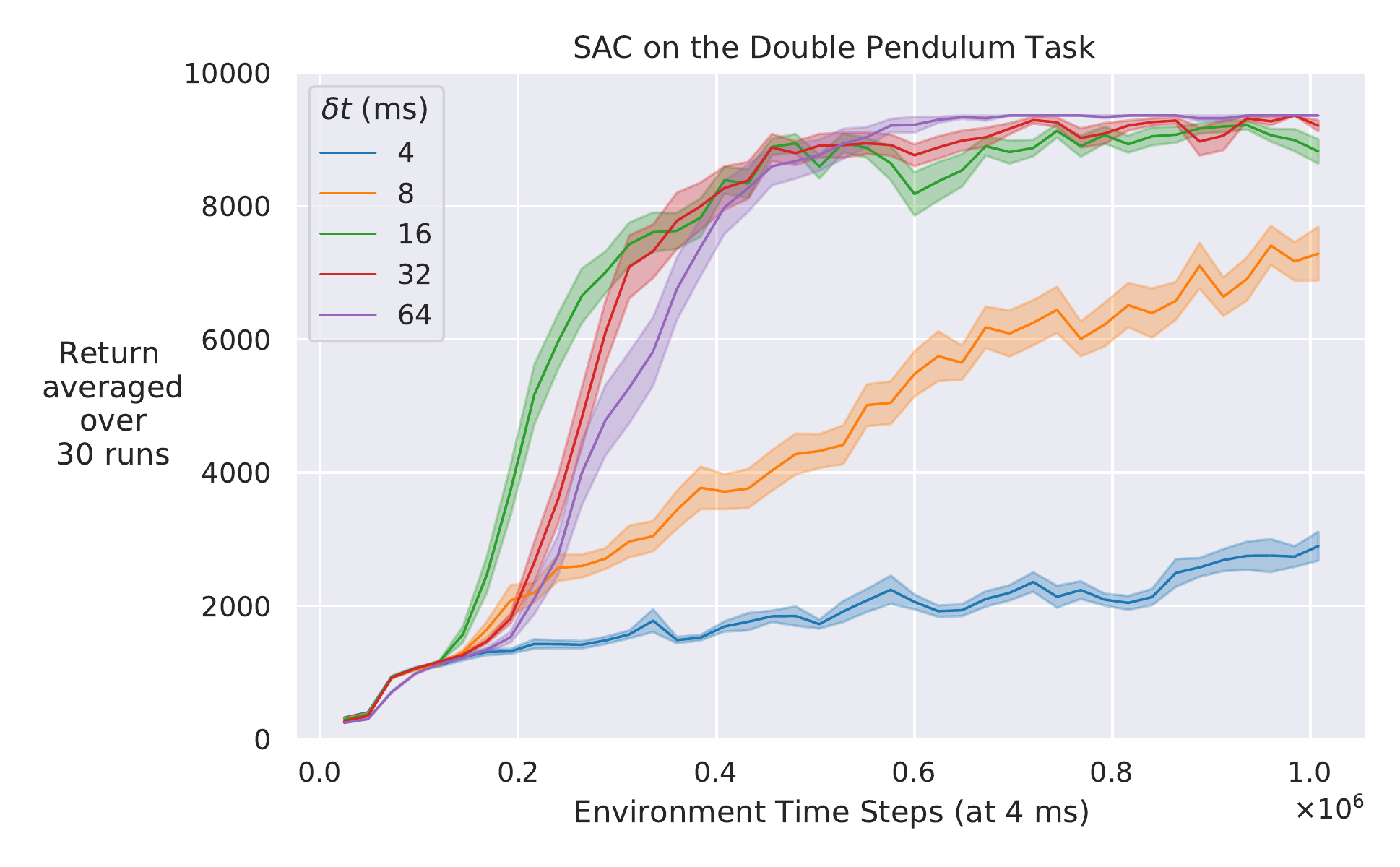}
        \caption[Learning curves of SAC using the baseline $\gamma$] {Corresponding learning curves of SAC for Figure \ref{fig:icra2022_oar_vs_dt_combo} (right) using the baseline $\gamma$. Learning speed of smaller \dt{}s is significantly impaired, and they seemingly could not reach their asymptote in the time given.}
        \label{fig:sac_oar_vs_dt_hyps_invdblpndlm_baseline_learning_curves}
    \end{minipage}\hfill
    \begin{minipage}[t]{.48\textwidth}
        \centering
        \includegraphics[keepaspectratio=true, width=1.0\textwidth]{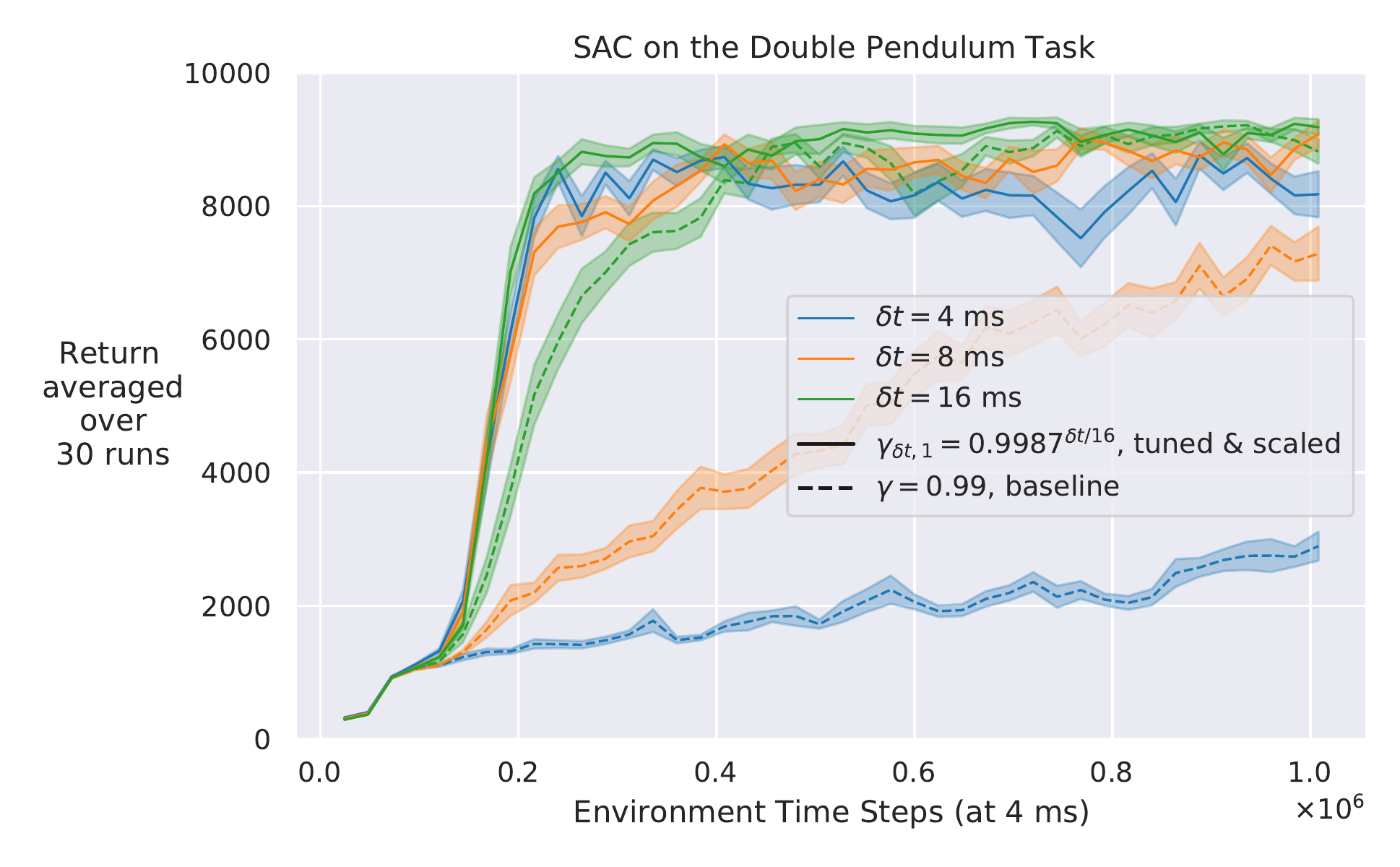}
        \caption[Learning curves of SAC comparing the baseline with the tuned and scaled $\gamma$] {Corresponding learning curves of SAC for Figure \ref{fig:icra2022_oar_vs_dt_combo} (right) comparing the baseline with the tuned and scaled $\gamma_{\delta t, 1}=0.9987^{\delta t/16}$. Performance of smaller \dt{}s is recovered. Larger \dt{}s were not drawn since no major change was observed compared to the baseline $\gamma$.}
        \label{fig:sac_oar_vs_dt_hyps_invdblpndlm_tunescale_learning_curves}
    \end{minipage}
\end{figure*}

\begin{figure*}[t]
    \centering
    \begin{minipage}[t]{.48\textwidth}
        \centering
        \includegraphics[keepaspectratio=true, width=1.0\textwidth]{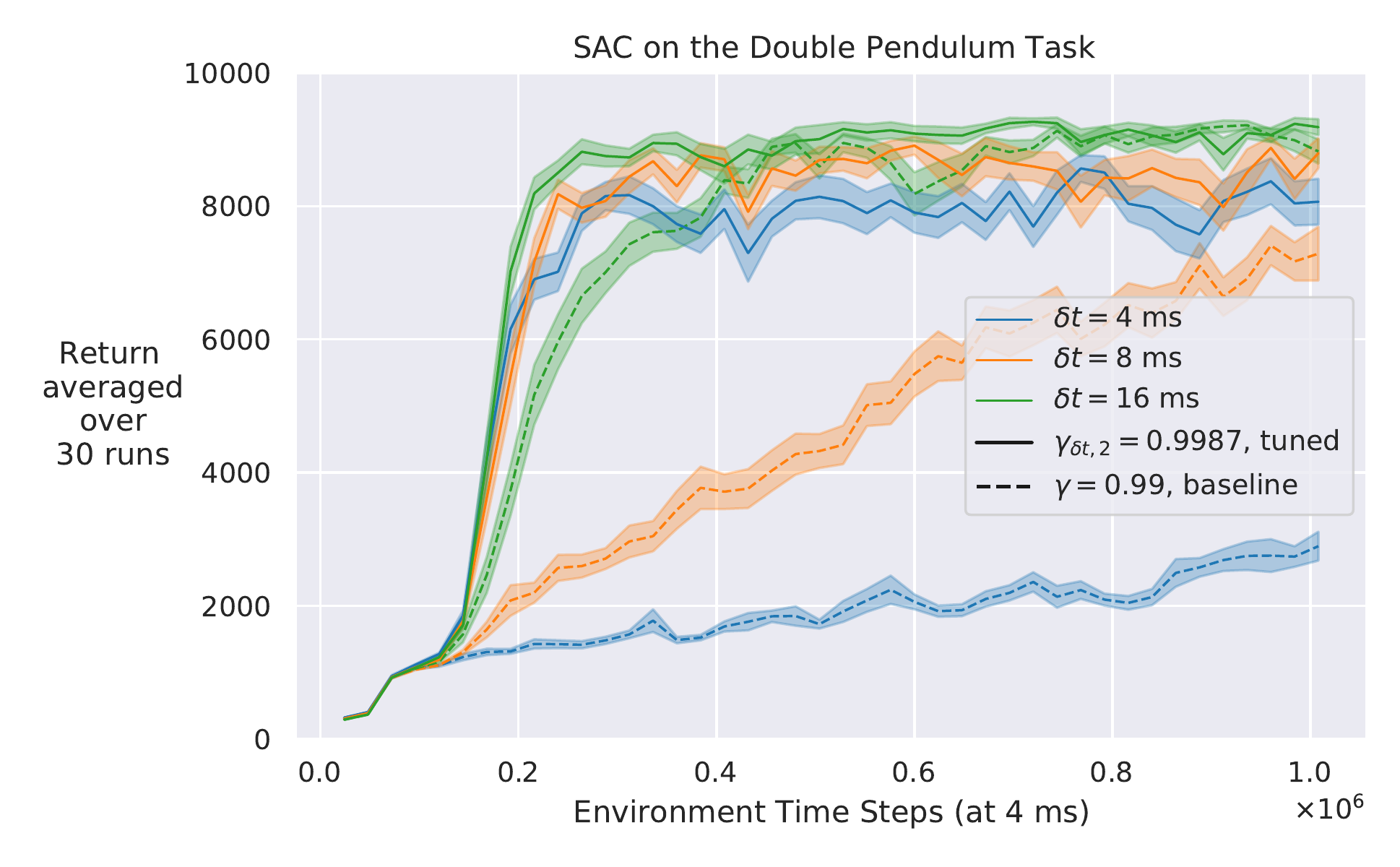}
        \caption[Learning curves of SAC comparing the baseline with the tuned $\gamma$] {Corresponding learning curves of SAC for Figure \ref{fig:icra2022_oar_vs_dt_combo} (right) comparing the baseline with the tuned $\gamma_{\delta t, 2}=0.9987$. Performance of smaller \dt{}s is recovered. Larger \dt{}s were not drawn since no major change was observed compared to the baseline $\gamma$.}
        \label{fig:sac_oar_vs_dt_hyps_invdblpndlm_tune_learning_curves}
    \end{minipage}\hfill
    \begin{minipage}[t]{.48\textwidth}
        \centering
        \includegraphics[keepaspectratio=true, width=1.0\textwidth]{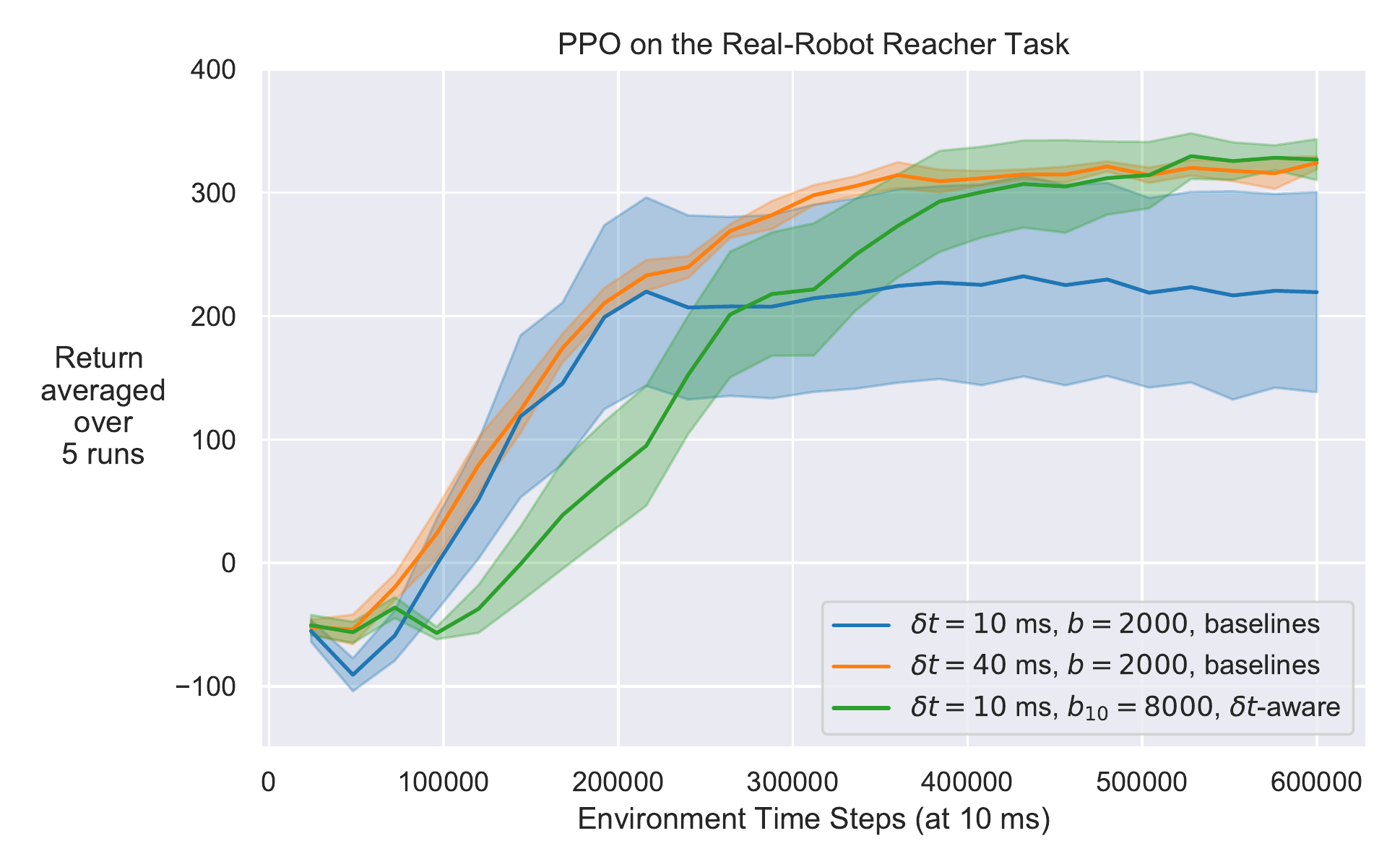}
        \caption[Learning curves of PPO on the Real-Robot Reacher Task using hyper-parameters of Table \ref{tbl:pposimhps_dtaware}] {Learning curves of PPO on the Real-Robot Reacher Task comparing the \delt{}-aware and baseline hyper-parameters of Table \ref{tbl:pposimhps_dtaware}. Performance of $\delta t=10$ms and ${\delta t}_0=40$ms are comparable with the baseline hyper-parameters.}
        \label{fig:ur5_learning_curves_2000}
    \end{minipage}
\end{figure*}

\begin{figure*}[t]
    \centering
    \begin{minipage}[t]{.48\textwidth}
        \centering
        \includegraphics[keepaspectratio=true, width=1.0\textwidth]{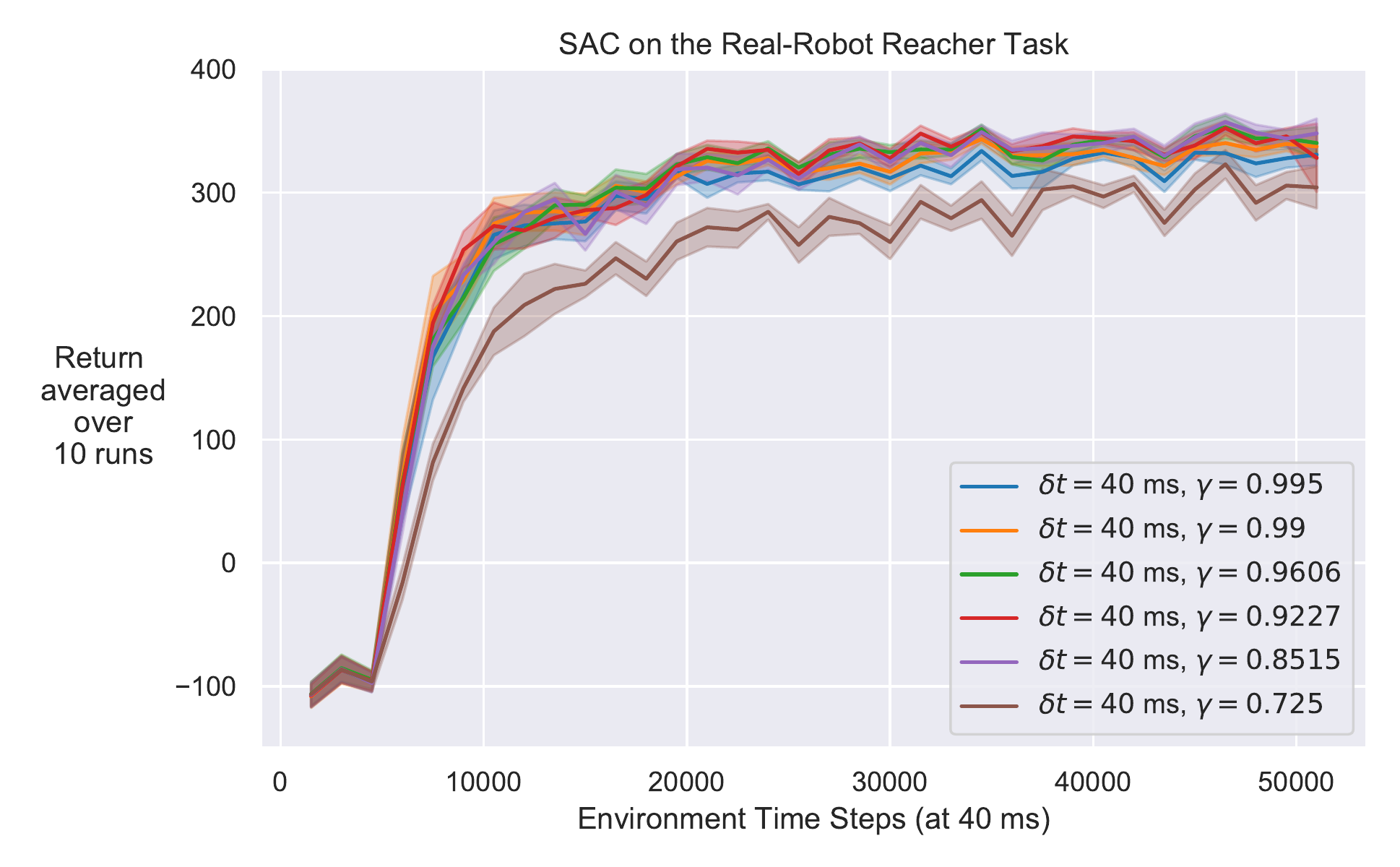}
        \caption[Learning curves of SAC on the Real-Robot Reacher Task for the sweep of $\gamma$ values at ${\delta t}_0=40$ms] {Corresponding learning curves of SAC for Figure \ref{fig:sac_oar_vs_gam_256_urreacher} on the Real-Robot Reacher Task for the sweep of $\gamma$ values at ${\delta t}_0=40$ms. The smallest $\gamma=0.725$ performs slightly worse than other $\gamma$s. Other $\gamma$s are roughly similar in learning speed and asymptotic performance.}
        \label{fig:sac_oar_vs_timesteps_256_urreacher_baseline_learning_curves}
    \end{minipage}\hfill
    \begin{minipage}[t]{.48\textwidth}
        \centering
        \includegraphics[keepaspectratio=true, width=1.0\textwidth]{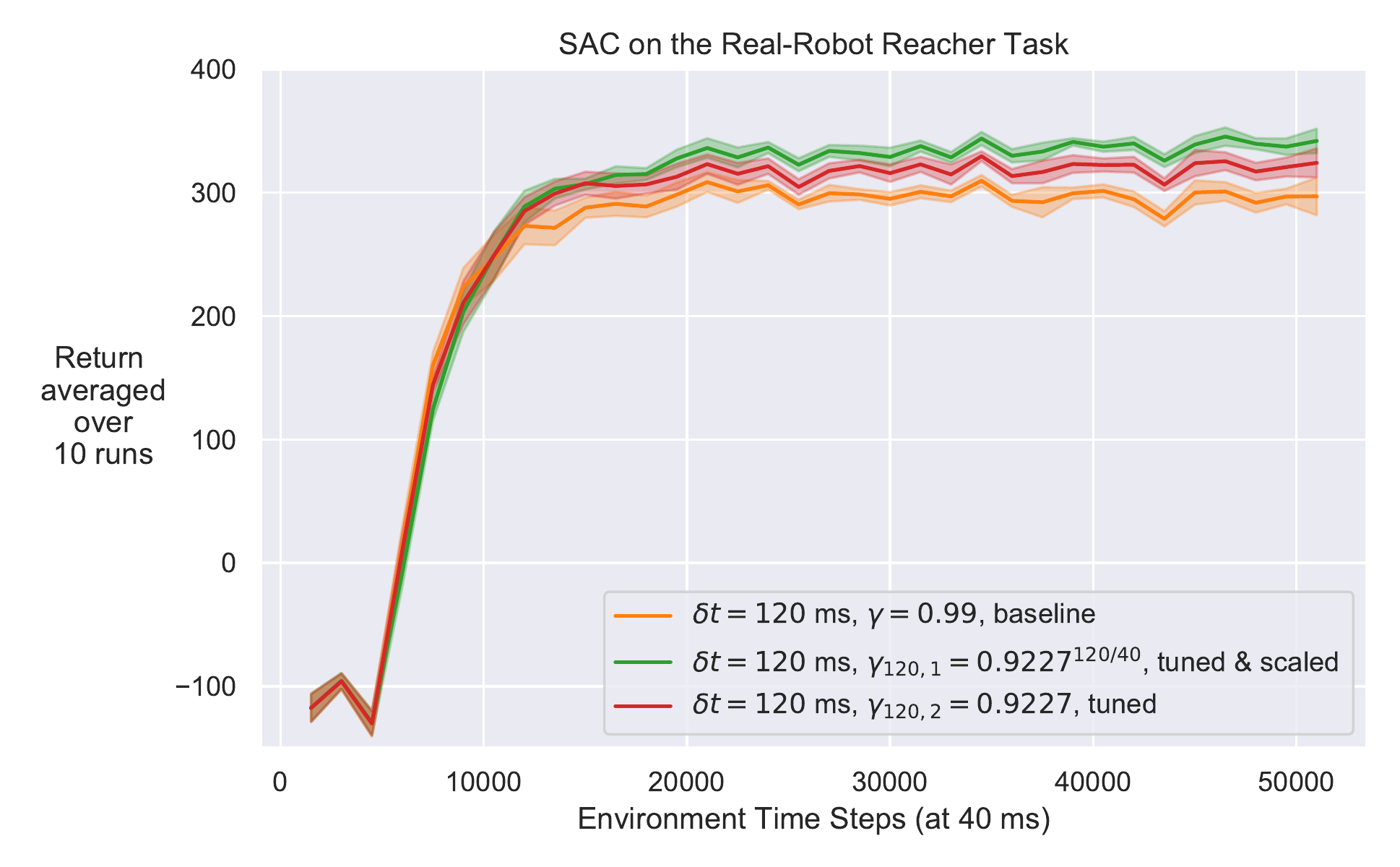}
        \caption[Learning curves of SAC on the Real-Robot Reacher Task at $\delta t=120$ms] {Corresponding learning curves of SAC for Figure \ref{fig:sac_oar_vs_gam_256_urreacher} on the Real-Robot Reacher Task for $\delta t=120$ms. The asymptotic performance of the baseline $\gamma=0.99$ cannot reach that of the atypical \delt{}-aware $\gamma_{\delta t, 1} \approx 0.786 $ even after extended learning.}
        \label{fig:sac_oar_vs_timesteps_256_urreacher_120ms}
    \end{minipage}
\end{figure*}

\end{document}